\documentclass[10pt,journal,compsoc]{IEEEtran}

\usepackage[nocompress]{cite}
\usepackage{hyperref}
\usepackage[pdftex]{graphicx}
\usepackage[cmex10]{amsmath}
\usepackage{amssymb}
\usepackage{algorithmic}
\usepackage[caption=false,font=normalsize,labelfont=sf,textfont=sf]{subfig}
\usepackage{multirow}
\usepackage{color}
\newcommand{\tabincell}[2]{\begin{tabular}{@{}#1@{}}#2\end{tabular}}

\usepackage{stfloats}

\begin{document}

\title{A Performance Evaluation of Local Features for Image-Based 3D Reconstruction}

\author{\IEEEauthorblockN{Bin Fan$^{1}$,  Qingqun Kong$^{1}$,  Xinchao Wang$^{2}$, Zhiheng Wang$^{3}$,  Shiming Xiang$^{1}$,  Chunhong Pan$^{1}$,  Pascal Fua$^{4}$}
\\
\IEEEauthorblockA{$^{1}$Institute of Automation, Chinese Academy of Sciences, China\\
$^{2}$Department of Computer Science, Stevens Institute of Technology, USA\\
$^{3}$School of Computer Science and Technique, Henan Polytechnic University, China\\
$^{4}$CVLab, EPFL, Switzerland\\
Email: bfan@nlpr.ia.ac.cn
}

}

\IEEEtitleabstractindextext{
\begin{abstract}
This paper performs a comprehensive and comparative evaluation of the state of the art local features for the task of image based 3D reconstruction. The evaluated local features cover the recently developed ones by using powerful machine learning techniques and the elaborately designed handcrafted features. To obtain a comprehensive evaluation, we choose to include both float type features and binary ones. Meanwhile, two kinds of datasets have been used in this evaluation. One is a dataset of many different scene types with groundtruth 3D points, containing images of different scenes captured at fixed positions, for quantitative performance evaluation of different local features in the controlled image capturing situations. The other dataset contains Internet scale image sets of several landmarks with a lot of unrelated images, which is used for qualitative performance evaluation of different local features in the free image collection situations. Our experimental results show that binary features are competent to reconstruct scenes from controlled image sequences with only a fraction of processing time compared to use float type features. However, for the case of large scale image set with many distracting images, float type features show a clear advantage over binary ones.
\end{abstract}
}

\maketitle

\IEEEdisplaynontitleabstractindextext

% \fi
\IEEEpeerreviewmaketitle

\IEEEraisesectionheading{\section{Introduction}}
At the core of image based 3D reconstruction systems~\cite{Agarwal_11,Heinly_CVPR15}, one fundamental task is to establish reliable point correspondences across multiple images of the reconstructed scene, which are captured from different viewpoints, positions, and usually at different time. Popular and still the dominant solution to this problem refers to matching keypoints by comparing their local descriptors. There are three typical steps involved in this procedure: extracting keypoints from images~(feature extraction), constructing local descriptors for keypoints~(feature description), and establishing point correspondences across different images according to distances of their descriptors~(feature matching).

In the past decade, various methods have been proposed to obtain keypoints and local descriptors as alternatives to the classical SIFT~\cite{LOWE_IJCV04} and SURF~\cite{Bay_CVIU08}. These methods either focus on the whole pipeline of feature extraction and description such as ORB~\cite{Rublee_ICCV11}, BRISK~\cite{Leutenegger_ICCV11}, FRIF~\cite{Wang_BMVC13}, KAZE~\cite{KAZE_ECCV12}, LIFT~\cite{LIFT_ECCV16}, or only focus on the descriptor, e.g., LIOP~\cite{Wang_ICCV11}, LDB~\cite{Yang_PAMI14}, VGGDesc~\cite{Simonyan_PAMI14}, BinBoost~\cite{Trzcinski_PAMI15}, DeepDesc~\cite{Edgar_ICCV15}, L2Net~\cite{L2Net_CVPR17}, and so on~\cite{Fan_TIP14,Wang_PAMI16,HardNet_NIPS17}. However, SIFT is still the major choice for the task of image based 3D reconstruction. Since all these follow-ups of SIFT have been claimed to outperform SIFT's performance on image matching and sometimes even with better computational efficiency~(for instance, in the case of binary features, ORB, BRISK, FRIF, etc.), it is straightforward to replace SIFT with these keypoints or local descriptors. What is the reason making the community does not to do so, at least up to nowadays? In this paper, we try to give an experimental study to answer this question. Specifically, we evaluate different combinations of keypoints and local descriptors for establishing point correspondences and embed matching points to an image based 3D reconstruction system. By doing so, we can obtain an end-to-end performance comparison of different keypoints and descriptors. Due to the large number of methods existed in this area, we choose to evaluate on the recent advances except for the classical SIFT, which is served as the baseline. To be specific, our evaluation covers both hand-crafted and learning based features with two different types: traditional float type ones and the emerging binary ones. For float type descriptors, it includes SIFT and LIOP as representative handcrafted ones and covers the learning based ones that use the traditional learning technique~(VGGDesc) and the recently popular CNNs~(DeepDesc, L2Net, LIFT). All these evaluated methods, except for SIFT and LIFT which have their own keypoint detectors, are merely feature description methods and so they have to be used with a keypoint detector. In this paper, we use SIFT keypoint for its popularity and also because that it is already used along with SIFT in the baseline. That is to say, except for LIFT, all the evaluated float type descriptors are based on SIFT keypoints, while LIFT is based on its own keypoints~(i.e., LIFT keypoints). For binary descriptors, we choose to evaluate the most recent ones, e.g., BRISK, FRIF, LDB, RFD~\cite{Fan_TIP14} and BinBoost. The former two are handcrafted features while the latter three are learned features. Among them, BRISK and FRIF contain both keypoint detector and binary descriptor. As a result, we use both of these two kinds of keypoints and combine them with all the evaluated binary descriptors respectively. It is worth to point out that although there are many works on local feature evaluation in the literature, most of them are limited to the image matching level~\cite{Mikolajczyk_PAMI05,Aanaes_IJCV12,Miksik_ICPR12,Heinly_ECCV12}.\\

For this comparative study, a basic but typical 3D reconstruction system is implemented\footnote{We will make our system and evaluation code public available.}. The system is based on the linear time incremental structure from motion~\cite{ChangChangWu_SFM_3DV13} and PMVS~\cite{Furukawa_PAMI10} by taking the matching keypoints across different images as input. We use different combinations of keypoint and local descriptor to generate different inputs to the system so as to obtain different reconstruction results. Two different types of datasets are used in our evaluation. The first one is a recently proposed multiview stereo dataset~(DTU MVS)~\cite{Jensen_CVPR14}, which contains more than 100 different scenes with high resolution images captured from 49 or 64 fixed viewpoints. Meanwhile, groundtruth 3D points are available. This dataset has a large diversity in scene types with a moderate number of images for each scene, while at the meantime still providing the groundtruth 3D points to facilitate an objective evaluation of reconstruction accuracy. The second dataset contains three large structure from motion~(SFM) subsets~\cite{Wilson_ECCV14}, which contains thousands of unordered images and many distracted images per scene. These two datasets stand for two typical image collection situations for 3D reconstruction applications. One is the controlled case where images are captured at selected viewpoints, and so is widely used for applications about reconstructing a very specific scene or object. In this case, all images cover a part of the scene and have moderate overlaps. The other case does not have any constraint on the used images, and so is widely used for applications about reconstructing a very large scale place such as a landmark or a city. In this case, it resorts to collect images from the Internet, instead of spending huge labors to capture high quality images with specially considered imaging viewpoints as in the first case. In this case, it inevitably contains many unrelated and low quality images as well as non-overlapping images, thus is more challenging.

The remaining parts of this paper are organized as follows. Section~\ref{sec:related_work} reviews the existing local features and their performance evaluations. In Section~\ref{sec:pipeline}, we briefly describe our implemented 3D reconstruction system. Then, the evaluated local features are introduced in Section~\ref{sec:float_feature} and Section~\ref{sec:binary_feature}. The evaluation results and analysis on the two used datasets are presented in Section~\ref{sec:MVS_result} and Section~\ref{sec:SFM_result} respectively. Finally, Section~\ref{sec:conclusion} concludes this paper.

\section{Related Work}
\label{sec:related_work}
\subsection{Keypoints and Local Image Descriptors}
Keypoint and local image descriptor are two critical parts of local feature. Keypoint detection aims to find re-detectable~(sparse) points in different images of a same scene. Such a re-detectable property, which is also known as the repeatability, is the principal consideration for designing a keypoint detector. In the literature, there are mainly two kinds of keypoints, corner points or blob points. Briefly speaking, they detect different types of image structures. For corner keypoints, the detectors seek local image structures that have large variance for different directions. Widely used methods include Harris~\cite{Harris_1988}, FAST~\cite{FAST_PAMI10} and AGAST~\cite{AGAST_ECCV10}. Due to the computational efficiency, FAST and AGAST have been used to detect scale invariant keypoints in scale spaces in recent binary features, e.g., ORB~\cite{Rublee_ICCV11} and BRISK~\cite{Leutenegger_ICCV11}. To detect blob-like image structures, the response of Laplacian of Gaussian~(LoG) filter~\cite{Lindeberg_IJCV98} and the determinant of Hessian matrix~\cite{Bay_CVIU08} are two widely used indicators. They are usually used in a scale space to search for local extrema so as to detect keypoints along with their characteristic scales. To accelerate the process of keypoint detection, several methods have been proposed to approximate the LoG detector, among which the most famous one is SIFT~\cite{LOWE_IJCV04}. CenSurE~\cite{Agrawal_ECCV08} approximated LoG by using the Bi-Level Laplacian of Gaussian. FRIF~\cite{Wang_BMVC13} proposed to use several box filters to approximate the LoG filter. Both CenSurE and FRIF use integral images for fast computation. In order to deal with severe viewpoint changes, local affine adaption techniques have been applied to keypoints in order to detect the so called interest regions, such as Hessian-Affine $\&$ Harris-Affine~\cite{Mikolajczyk_IJCV04}. Another well known interest region is MSER~\cite{MSER_BMVC02}, which detects stable gray-scale regions. Since MSER could be with any shape, it is usually to fit MSER by ellipse based on second order moments. While all these methods are build up on a formal definition of keypoints, some other works leverage on labeled data to learn keypoint detectors, such as~\cite{TILDE_CVPR15,LIFT_ECCV16,Zhang_CVPR17}.

To match keypoints or interest regions, the common practice is to construct a local image descriptor for each of them, and then build correspondence between them based on the descriptors' distances. For this purpose, a local image descriptor is expected to be designed with high robustness in order to tolerate with various photometric or geometric transformations among the corresponding local regions. In this way, keypoints corresponding to the same physical points can be correctly matched. At the meantime, it is also expected to be with high distinctiveness so that keypoints corresponding to different physical points can be easily distinguished. The community has made great efforts to achieve these two goals simultaneously. The milestone work is no doubt the SIFT~\cite{LOWE_IJCV04}, after which many handcrafted local descriptors have been proposed, such as SURF~\cite{Bay_CVIU08}, LIOP~\cite{Wang_ICCV11}, KAZE~\cite{KAZE_ECCV12}, and so on~\cite{Marko_PR09,Fan_PAMI12,Wang_PAMI16}. All these descriptors were reported with better performance than SIFT in some aspects, for example, dealing with complex brightness changes or image blur, or computational efficiency. With the access of a huge number of matching and non-matching local patches~\cite{Matthew_PAMI11}, researchers have gradually moved their interest from handcrafted methods to the learning based ones. Matthew et al.~\cite{Matthew_PAMI11} proposed to learn discriminative local descriptors by optimizing over the combination of low-level features and spatial pooling methods as well as their parameters. The dimension of the learned descriptors can be further reduced to very small by applying subspace embedding. Following this work, Simonyan et al.~\cite{Simonyan_PAMI14} reformulated the learning problem as a sparse constrained convex optimization problem. Recently, deep learning has been applied to learn high matching performance local descriptors. Han et al.~\cite{MatchNet_CVPR15} proposed the MatchNet to unify descriptor learning and metric learning in a framework by maximizing the descriptor distance between non-matching patches and minimizing that of matching patches. MatchNet not only learns the patch descriptor, but also their distance metric. Similar learning paradigm has been used by Zagoruyko and Komodakis~\cite{DeepCompare_CVPR15} and Kumar et al.~\cite{Kumar_CVPR16}. Although these methods lead to high matching performance, they have to be used along with the learned metric. Using the learned descriptor alone can not guarantee the good matching performance. Such a constraint largely limits their applications and a drop-in replacement of the previous handcrafted descriptors is highly required. For this purpose, learning patch descriptor that can be directly matched in the Euclidean space has received great interest in the recent two years. The representative works include DeepDesc~\cite{Edgar_ICCV15}, TFeat~\cite{TFeat_BMVC16}, L2Net~\cite{L2Net_CVPR17} and HardNet~\cite{HardNet_NIPS17}.

\subsection{Performance Evaluation of Local Features}
Accompany with the flourish of local features, many works have been conducted to evaluate performance of various local features under the scope of different applications. Mikolajczyk et al.~\cite{Mikolajczyk_PAMI05,Mikolajczyk_IJCV05} evaluated the matching performance of different local descriptors and affine invariant interest regions in the task of matching images of planar scenes. Moreels and Perona~\cite{Moreels_IJCV07} extended Mikolajczyk's evaluations to images of 3D objects captured on a turntable. These evaluations demonstrated the higher distinctiveness of SIFT than its previous methods, thus promoting the development of SIFT-like local features, i.e., histogram-based handcrafted features such as SURF~\cite{Bay_CVIU08}, DAISY~\cite{DAISY_PAMI10}, CS-LBP~\cite{Marko_PR09}, KAZE~\cite{KAZE_ECCV12}. Aan{\ae}s et al.~\cite{Aanaes_IJCV12} revised Mikolajczyk's and Moreels's works by introducing a more comprehensive dataset with known spatial correspondence of points, while at the meantime to cover various situations for interest point matching. Although most detectors in their evaluation has been evaluated before~\cite{Mikolajczyk_PAMI05,Moreels_IJCV07}, their evaluation was more thorough and convincing because the newly introduced dataset is more realistically challenging. Their evaluation re-emphasized the importance of detecting feature points in scale space and showed that the affine adaption proposed by Mikolajczyk and Schmid~\cite{Mikolajczyk_IJCV04} has a little influence on feature detector itself, but is useful for the descriptor, thus is helpful in the whole pipeline of feature matching. Recently, with the development of binary descriptors, some researchers evaluated different local features under the same evaluation protocol of image matching as~\cite{Mikolajczyk_PAMI05} but with an emphasize on the compactness and speed of the tested methods. For this purpose, Miksik and Mikolajczyk~\cite{Miksik_ICPR12} shown that binary features such as ORB~\cite{Rublee_ICCV11} and BRIEF~\cite{Calonder_PAMI11} are efficient for both feature extraction and matching for image matching due to the fast computation of Hamming distance. On the other hand, state of the art handcrafted descriptors such as LIOP~\cite{Wang_ICCV11} and MROGH~\cite{Fan_PAMI12} could result in better matching performance but with much higher computational burden. Similarly, Heinly et al.~\cite{Heinly_ECCV12} gave a comparative evaluation of binary features by considering not only the classical performance metrics such as precision and recall, but also introducing new metrics such as the spatial distribution of the features as well as the frequency of candidate matches.

All the above evaluations were conducted for the task of image matching. For other applications, Gauglitz et al.~\cite{Gauglitz_IJCV11} evaluated different interest points and local descriptors for visual tracking. Bauml and Stiefelhagen~\cite{Bauml_AVSS11} evaluated different local features for person re-identification in image sequences. Madeo and Bober~\cite{Madeo_TMM17} conducted a comparative study on using binary descriptors for mobile applications. Liu et al.~\cite{Liu_ECCV16,Liu_PR17} conducted evaluations of local binary features for texture classification. Similar to this paper, Fan et al.~\cite{Fan_CVPRW16} and Schonberger et al.~\cite{Schonberger_CVPR17} studied performance of different local features for image based 3D reconstruction systems. However, Fan et al.~\cite{Fan_CVPRW16} only evaluated three binary features~(ORB, BRISK and FRIF) that contain both feature detector and descriptor while Schonberger et al.~\cite{Schonberger_CVPR17} were mainly focused on the learned float type descriptors. On the contrary, this paper extensively evaluates different combinations of existing binary descriptors and feature detectors. Besides traditional handcrafted ones, these binary descriptors also include learning based ones, e.g., BinBoost~\cite{Trzcinski_PAMI15}, LDB~\cite{Yang_PAMI14} and RFD~\cite{Fan_TIP14}, which have been shown with superior performance on standard image matching benchmarks. Moreover, a comparative study of the state of the art float type descriptors is conducted in this work too. Therefore, the evaluation of this work is more comprehensive compared to the previous works, covering the state of the arts in both binary and float type local features, and ranging from handcrafted features to the learning based ones. Many of these features are not evaluated before. What is more, about the evaluation datasets, we use both the DTU MVS dataset~\cite{Jensen_CVPR14} used in Fan et al.~\cite{Fan_CVPRW16} and the large scale SFM dataset~\cite{Wilson_ECCV14} used in Schonberger et al.~\cite{Schonberger_CVPR17}. In this way, our evaluation covers two typical cases for 3D reconstruction, i.e., 1) controlled image capturing with moderate number of images, and 2) free image capturing with a large number of images and many distracted images. For the former case, we rely on the supplied groundtruth to study the performance~(accuracy and completeness of the reconstruction) of different feature combinations. While for the latter, the ability of reconstructing scene from as many images as possible is what we pursue.

\section{Pipeline of 3D Reconstruction}
\label{sec:pipeline}
To obtain the 3D points of an object or a scene by only using a number of images, the popular solutions~\cite{Agarwal_11,Heinly_CVPR15} usually include three steps: feature matching across images, structure from motion~\cite{MRF_SFM_PAMI13,ChangChangWu_SFM_3DV13} and dense reconstruction~\cite{Furukawa_PAMI10}. Feature matching aims to find the so called feature tracks. In essential, a feature track corresponds to a 3D point, containing point correspondences across different images. For unordered and very large scale image collection, there is usually an additional preprocessing step, aiming to quickly find out possible overlapping image pairs so as to conduct feature matching only on these pairs to save matching time~\cite{Lou_ECCV12,Schonberger_CVPR15_1}. Structure from motion takes a number of feature tracks as input, and outputs a number of 3D points as well as some camera parameters of the input images. With the recovered cameras, dense reconstruction is applied to obtain a dense 3D point cloud as the reconstruction result. In a word, a typical 3D reconstruction system outputs include a number of 3D points of the scene and the estimated camera parameters of the input images. By comparing these outputs to the groundtruth, one can evaluate how good the system is, e.g., in terms of 3D reconstruction accuracy, completeness and successfully recovered cameras.

In this paper, we focus on the step of feature matching, studying its performance when using different local features. As a result, we fix the last two steps with typical methods: linear time incremental structure from motion~\cite{ChangChangWu_SFM_3DV13} and PMVS~\cite{Furukawa_PAMI10} for dense reconstruction. Their source codes are provided and can be downloaded from their websites. Meanwhile, no preprocessing is used, i.e., feature matching is extensively conducted for all possible image pairs. In the following, we give a brief introduction to the evaluated features first and then move to the evaluation.

\section{Float Type Features}
\label{sec:float_feature}
Local feature has been an active and persistent topic in computer vision community. To keep this evaluation thorough and up to data, we choose recently proposed methods, including both handcrafted descriptors and the recent popular learning based ones. For reference, we also include the classical SIFT in our evaluation as baseline.

\subsection{SIFT}
SIFT constructs a Difference of Gaussian~(DoG) scale space to detect extrema across both spatial and scale spaces as keypoints. DoG scale space is constructed by subtracting neighboring images of a Gaussian scale space of the input image. The keypoint orientation is computed by accumulating a histogram of gradient orientations from a local circular region around the keypoint to achieve rotation invariance. The orientation corresponding to the largest bin in this histogram is taken as the keypoint orientation. Meanwhile, other orientations corresponding to the peak bins which are within 80\% of the largest one are also taken as the keypoint's orientations.

For feature description, SIFT divides the scale and rotation normalized local patch around a keypoint into $4 \times 4$ grids. In each grid, it computes a histogram of gradient orientations with 8 bins. All these histograms are concatenated together and normalized to get a 128 dimensional float vector as the SIFT descriptor. To improve robustness, the trilinear interpolation among spatial and orientation bins is utilized and a Gaussian weight is assigned to each pixel in the local patch.

\subsection{LIOP}
In SIFT and and its variants~\cite{Mikolajczyk_PAMI05,Bay_CVIU08,KAZE_ECCV12,Marko_PR09}, they rely on dominant orientations to achieve rotation invariance. Fan et al.~\cite{Fan_PAMI12} observed that the dominant orientations estimated from local image context are unreliable, and thus they proposed to construct local image descriptors by intensity order pooling to achieve intrinsic rotation invariance. Under this framework, Wang et al.~\cite{Wang_ICCV11} proposed the LIOP descriptor by pooling a kind of low level feature based on the local ordinal information around a pixel in the support region. The local intensity order can explore the relative relationship of intensities among all neighboring points around a pixel, not merely the relationship between two points which is often used by LBP invariants~\cite{Liu_TIP16,Liu_InforSci16,Chen_BMVC13,Chen_TIP13}. As a result, LIOP was reported with higher performance than its previous methods. For this reason, we choose to include LIOP in our evaluation as a representative handcrafted local feature.

\subsection{VGGDesc}
While traditional methods for local image description are handcrafted, learning good local descriptors has been extensively explored in recent years. One representative work of this type is proposed by the Visual Geometry Group~(VGG) in the Oxford University. Following Brown et al.'s work on discriminative learning of local image descriptors~\cite{Matthew_PAMI11}, Simonyan et al.~\cite{Simonyan_PAMI14} proposed to formulate the descriptor learning problem in a convex optimization framework based on the hinge loss with sparsity constraint. They used the RDA~\cite{RDA_JMLR10} to efficiently solve the involved sparse constrained optimization problem with large scale training set. They first learned a high dimensional descriptor by selecting discriminative pooling areas through sparse constraint. Then, they pursued a linear subspace of the learned high dimensional descriptor to obtain the final compact descriptor with powerful discriminative ability.

\subsection{DeepDesc}
With the popularity of using Convolutional Neural Networks~(CNNs) in various vision tasks, it has also been used in descriptor learning. Although initial works on using CNNs to learn patch descriptors are usually combined with additional metric layers to achieve good matching performance~\cite{MatchNet_CVPR15,DeepCompare_CVPR15,Kumar_CVPR16}, researchers gradually move to the more practical case, i.e., learning a patch descriptor than can be directly operated in the Euclidean space. This is because that this kind of descriptor can be used as a drop-in replacement for the widely used handcrafted descriptors, thus has wider applications. One representative work of learning patch descriptors without additional metric layers is the DeepDesc proposed by Edgar et al.~\cite{Edgar_ICCV15}. They used a Siamese Network structure and minimized a hinge-like loss when training the network. With a carefully designed network structure and a hard sample mining strategy for network training, they finally obtained a 128 dimensional float type descriptor that can be measured in the Euclidean space.

\subsection{L2Net}
A very recent work on learning discriminative patch descriptor in the Euclidean space by CNNs is the L2Net~\cite{L2Net_CVPR17}, which is specially designed for the matching task and incorporates supervision information of intermediate layers to improve its generalization ability. It takes a fully convolutional architecture with 7 convolutional layers, each of which is followed by a batch normalization layer with fixed parameters. Like DeepDesc, it finally outputs a 128 dimensional vector as the descriptor to serve as a drop-in replacement of SIFT for various applications. L2Net was the rank one method for the competition of local features held in ECCV'16 and obtained the top performance on the widely used patch matching dataset~(i.e., the Brown dataset~\cite{Matthew_PAMI11}). Due to its superior performance, we choose to include it in our evaluation.

\subsection{LIFT}
We also include LIFT~\cite{LIFT_ECCV16} in this evaluation as the state of the art method for the whole pipeline of feature detection and description. Inspired by the success of deep learning and identical to the SIFT's pipeline, LIFT combines all necessary components~(i.e., keypoint detector, orientation estimator, and local patch descriptor) of a local feature altogether in an end-to-end manner based on the deep convolutional architecture. Specifically, it uses TILDE~\cite{TILDE_CVPR15} as the keypoint detector because TILDE is convolutional, differentiable and with good performance. After detecting keypoints, it estimates the orientations of those patches around the detected keypoints by a CNN which is trained to minimize the generated descriptors' distance of matching patches~\cite{Yi_CVPR15}. Finally, the DeepDesc is used to extract feature descriptors for the scale and rotation normalized patches. To crop, resize, and rotate the local patch around a keypoint, LIFT uses the spatial transform network~\cite{SPN_NIPS15} as connector since it is differentiable. As a result, the whole pipeline of LIFT is differentiable and so can be trained in an end-to-end manner. In practice, the authors trained LIFT one component by one component started from the descriptor part and then finetuned the whole pipeline.

\subsection{Implementation Details}
For SIFT, we use the implementation supplied in VLFeat~\cite{vlfeat}. For the other float type descriptors, we use the implementations provided by their authors\footnote{\scriptsize{
LIFT: \href{https://github.com/cvlab-epfl/LIFT}{https://github.com/cvlab-epfl/LIFT}

LIOP: \href{https://github.com/foelin/IntensityOrderFeature}{https://github.com/foelin/IntensityOrderFeature}

L2Net: \href{https://github.com/yuruntian/L2-Net}{https://github.com/yuruntian/L2-Net}

DeepDesc: \href{https://github.com/etrulls/deepdesc-release}{https://github.com/etrulls/deepdesc-release}

VGGDesc: \href{http://www.robots.ox.ac.uk/~vgg/software/learn_desc/}{http://www.robots.ox.ac.uk/~vgg/software/learn\_desc/}
}}. SIFT keypoints~(i.e., DoG) are used for all these descriptors except for LIFT, which has its own keypoints. The low dimensional descriptor learned on the 'Liberty' of the Pacth Dataset~\cite{Matthew_PAMI11} is used for the VGG descriptor. Similarly, the evaluated L2Net is also trained on the 'Liberty'. While for the DeepDesc, we use the authors' suggested model that was trained on a subset of 'Liberty', 'Notre Dame' and 'Yosemite' of the Patch Dataset. For LIFT, it wa trained with a SFM dataset~(Piccadilly Circus dataset~\cite{Wilson_ECCV14}), and we use the public available model supplied by the authors. Please see Table~\ref{tab:feature} for a summary of all these local features. Identical to the Lowe's ratio test~\cite{LOWE_IJCV04}, the Nearest Neighbor Distance Ratio~(NNDR) is used for matching keypoints, where the ratio threshold is set as 0.8 for all the tested descriptors. To find the nearest and the second nearest neighbors, we use the open source ANN library~\cite{ANN} for the fast approximate nearest neighbor search.

\begin{table*}[!htb]
\renewcommand{\arraystretch}{1.1}
\centering
\begin{tabular}{|c|c|c|c|c|c|c|}
\hline
keypoint & descriptor & dimension & data type & handcrafted & learned & training set \\
\hline
\multirow{5}{*}{FRIF or BRISK}  &  FRIF~\cite{Wang_BMVC13} & 512 & binary & $\surd$  & $\times$ & $\times$  \\
\cline{2-7}
     &    BRISK~\cite{Leutenegger_ICCV11} & 512 & binary & $\surd$ & $\times$ & $\times$ \\
\cline{2-7}
     &    LDB~\cite{Yang_PAMI14} & 256 & binary & $\times$ & $\surd$ & Liberty~\cite{Matthew_PAMI11} \\
\cline{2-7}
     &    RFD~\cite{Fan_TIP14} & 288 & binary & $\times$ & $\surd$ & Liberty~\cite{Matthew_PAMI11} \\
\cline{2-7}
     &    BinBoost~\cite{Trzcinski_PAMI15} & 256 & binary & $\times$ & $\surd$ & Liberty~\cite{Matthew_PAMI11} \\
\hline
\multirow{5}{*}{DoG~(SIFT)}  &  SIFT~\cite{LOWE_IJCV04} & 128 & float & $\surd$  & $\times$ & $\times$  \\
\cline{2-7}
     &    LIOP~\cite{Wang_ICCV11} & 144 & float & $\surd$ & $\times$ & $\times$ \\
\cline{2-7}
     &    VGGDesc~\cite{Simonyan_PAMI14} & 128 & float & $\times$ & $\surd$ & Liberty~\cite{Matthew_PAMI11} \\
\cline{2-7}
     &    DeepDesc~\cite{Edgar_ICCV15} & 77 & float & $\times$ & $\surd$ & \tabincell{c}{subset of \\\{Liberty,NotreDame,Yosemite\}~\cite{Matthew_PAMI11}}  \\
\cline{2-7}
     &    L2Net~\cite{L2Net_CVPR17} & 128 & float & $\times$ & $\surd$ & Liberty~\cite{Matthew_PAMI11} \\
\hline
  LIFT   &    LIFT~\cite{LIFT_ECCV16} & 128 & float & $\times$ & $\surd$ & Piccadilly~\cite{Wilson_ECCV14} \\
\hline
\end{tabular}
\caption{Summary of the evaluated local features.
\label{tab:feature}}
\end{table*}

\section{Binary Features}
\label{sec:binary_feature}
To reduce the memory footprint of float type descriptors, binary descriptors have been widely studied in recent years. These binary descriptors have been used in some light weight tasks, such as template based object detection~\cite{Calonder_PAMI11} and SLAM~\cite{ORB_SLAM_TR15}, which usually involve matching only several hundreds of keypoints. However, they have not yet been used or evaluated for tasks involving extensively keypoint matching, such as the one we studied in this paper. In this work, we choose typical binary features to evaluate their performance on 3D reconstruction. For comprehensiveness, we cover both handcrafted ones and the learning based ones as summarized on Table~\ref{tab:feature}.

\subsection{BRISK}
BRISK contains a scale and rotation invariant keypoint detector and a binary feature descriptor. For the keypoint detector, BRISK implements a scale space by using two pyramids alternately, one for the octaves and the other for the intra-octaves, to trade-off the computation and scale estimation accuracy. The keypoints are detected in each level of the scale space based on the AGAST~\cite{AGAST_ECCV10}, which is an effective extension of the FAST corner detector~\cite{FAST_PAMI10}. Based on the position and scale of the detected keypoint, a sampling pattern with 60 points regularly sampled from 4 concentric circles are used to compute the keypoint's orientation as well as its binary descriptor. Specifically, the point pairs generated by these sampling points are divided into long-distance pairs and short-distance ones. The long-distance pairs are used to compute an average local gradient to define the orientation of the keypoint, while the short-distance pairs are used for intensity tests to construct the binary descriptor. To deal with aliasing effects, the intensity of a sampling point is computed by filtering with a Gaussian kernel whose standard deviation is proportional to its distance to the keypoint, i.e., the central point of the sampling pattern.

\subsection{FRIF}
While BRISK resorts to FAST detector for efficient keypoint detection, FRIF relies on the response of Laplacian of Gaussian~(LoG). The basic idea is to approximate LoG with rectangular filters so that to compute its response very quickly by integral images. According to Mikolajczyk and Schmid's study~\cite{Mikolajczyk_ICCV01}, Laplacian of Gaussian is stable in characteristic scale selection and has been used in many feature detectors~\cite{Mikolajczyk_IJCV04,LOWE_IJCV04}. In FRIF, it approximates a LoG template by linear combination of four rectangular filters. Therefore, computing the LoG responses on pixels of an image just requires linear combination of four rectangular filtering results, which can be done efficiently based on integral images. To detect extrema of the approximated LoG responses across both spatial and scale spaces, FRIF implements an identical scale space as BRISK does and uses a similar strategy for non-maximum suppression as well as location refinement.

As far as the binary descriptor is concerned, FRIF uses a similar sampling pattern to BRISK, but proposes a mixed binary descriptor to achieve better performance. For each sampling point, it uses its neighboring points to conduct intensity tests to obtain a number of bits as part of the descriptor. It also uses some short-distance point pairs for intensity tests as the remaining part of the descriptor to capture complementary information. The long-distance point pairs are used to compute the keypoint orientation as in BRISK.

\subsection{LDB}
LDB~\cite{Yang_PAMI14} is a binary descriptor computed based on intensity difference and gradient difference. It first participates the local region into several cells according to the predefined spatial configurations. Then the averaged intensities and gradients are computed for each of these cells. These average values
between cell pairs are compared to generate binary values so as to construct the binary descriptor. To select only a few discriminative and meaningful test pairs from all the possible cell pairs, a modified adaboost algorithm is proposed by Yang and Cheng~\cite{Yang_PAMI14}.

\subsection{RFD}
Gradient orientation map used in SIFT and DAISY~\cite{DAISY_PAMI10} has shown its effectiveness in constructing discriminative local descriptors. Fan et al.~\cite{Fan_TIP14} extended it for binary feature description. They proposed to construct a bit of a binary descriptor by thresholding the oriented gradient responses accumulated from a certain region, which is either a rectangular or a Gaussian shaped region. The best threshold value for each region is determined by the Bayesian criteria according to the labeled training data. Such regions constructing the so call RFD descriptor are greedy selected from a large pool of candidates according to their discriminative ability and correlation.

\subsection{BinBoost}
Similar to RFD which uses the thresholded gradient orientation map as the basic element, Trzcinski et al.~\cite{Trzcinski_PAMI15} applied boosting to learn high compact binary descriptor. The learned descriptor, named as BinBoost, takes a linear combination of several thresholded gradient orientation maps and then thresholds the combination result as one bit in the descriptor. In other words, if we consider each gradient orientation map as one weak classifier, each bit in BinBoost corresponds to a strong classifier according to the boosting theory. The gradient orientation maps and their linear weights are selected based on a modified adaboost learning algorithm proposed in their paper too.

Among the above five binary descriptors, the first two have both feature detector and feature descriptor. The latter three are only binary descriptors which have to be evaluated along with a specific feature detector. Therefore, in our evaluation, we combine them with feature detectors provided by the first two methods respectively. Here, we do not evaluate ORB~\cite{Rublee_ICCV11} for two reasons. First, both BRISK keypoint and ORB keypoint are based on the AGAST while BRISK uses a finer scale space, so the BRISK keypoint is better. Second, ORB has been shown with inferior performance to BRISK and FRIF in our previous work~\cite{Fan_CVPRW16}.

\subsection{Implementation Details}
All the evaluated binary features have source codes available on the Internet, therefore, we use the original implementations with default parameters released by their authors\footnote{\scriptsize{
RFD: \href{http://www.nlpr.ia.ac.cn/fanbin/rfd.htm}{http://www.nlpr.ia.ac.cn/fanbin/rfd.htm}

LDB: \href{http://lbmedia.ece.ucsb.edu/research/binaryDescriptor/web_home/web_home/index.html}{http://lbmedia.ece.ucsb.edu/research/binaryDescriptor/web\_home/web\_home}

FRIF: \href{https://github.com/foelin/FRIF}{https://github.com/foelin/FRIF}

BRISK: \href{http://www.asl.ethz.ch/people/lestefan/personal/BRISK}{http://www.asl.ethz.ch/people/lestefan/personal/BRISK}

BinBoost: \href{http://cvlab.epfl.ch/research/detect/binboost}{http://cvlab.epfl.ch/research/detect/binboost}
}}. For RFD, the one trained on the 'Liberty' of the Patch Dataset with rectangle receptive field is used~(denoted as RFDR). For BinBoost, the one with 256 bits is used, which is also trained on the 'Liberty' and reported with the best generalization ability.

To match keypoints of these binary features, we use the multi-table and multi-probe LSH implemented in the FLANN library~\cite{FLANN} to approximately find the first two nearest neighbors in an efficient manner. Then the distance ratio of the first and the second nearest neighbors is used to decide whether two keypoints are matched or not. The same as the case of float descriptor, the ratio threshold is set as 0.8. Note that although computing the Hamming distance of two binary descriptors is significantly faster than computing the Euclidean distance of two float type descriptors, it is still impractical to conduct bruteforce nearest neighbor search in Hamming space because of the large number of image matching operations involved in 3D reconstruction task. Due to this reason, the fast approximate nearest neighbor search method, i.e. multi-table, multi-probe LSH, is used. Specifically, we set the number of hash tables as 4, the multi-probe level as 1, the LSH code length as 24 in all our evaluations.

\section{Evaluation on Multiview Stereo Dataset}
\label{sec:MVS_result}

\subsection{Dataset}

We first choose to evaluate the 3D reconstruction performance of different features on a recently published multiview stereo dataset, known as the DTU MVS dataset~\cite{Jensen_CVPR14}. It contains a total number of 124 different scenes, covering a wide range of objects and surface materials. For each scene, it collects images of $1600 \times 1200$ resolution from 49 or 64 different viewpoints, with 8 different illumination conditions. Among these scenes, 80 scenes contain necessary information~(i.e., observability mask) that is required for the evaluation of reconstruction results as Jensen et al. did~\cite{Jensen_CVPR14}. In this paper, we use the scenes with 49 views, which occupy 58 out of all 80 scenes. We do not study effects of different lighting conditions, so we just use the subset with all lights on.

Due to the fact that our implemented 3D reconstruction system is fully automatic and uses the self-calibration to decide the camera parameters, the coordinate system of the reconstructed 3D points can be any of those recovered cameras. In this case, the reconstructed coordinate system and the supplied reference coordinate system are related by a 3D similarity transformation~(scaling, rotation and translation). Therefore, we have to firstly register the reconstructed 3D points to the reference scans~(groundtruth) obtained by a structure light scanner which are supplied in the dataset. To this end, we manually selected three corresponding 3D points between the reconstructed one and the groundtruth. Then, they are used to estimate a similarity transformation to register the reconstructed 3D points.

\subsection{Evaluation Protocol}

After registering the reconstructed 3D points to the reference coordinate system, we use the supplied code in the dataset for performance evaluation. The evaluation protocol is based on that of~\cite{Seitz_CVPR06}, with some modifications to make it unbiased and better at handling missing data and outliers. Basically, it adopts an observability mask so that the evaluation is only focused on the visible part of the scene. Please refer to~\cite{Jensen_CVPR14} for more details.

As in~\cite{Seitz_CVPR06,Jensen_CVPR14}, accuracy and completeness are used as quality measures of a reconstruction. According to their definitions, given a reconstruction and the structure light reference, the accuracy is computed as the distance from the reconstruction to the reference scan. On the contrary, the completeness is computed as the distance from the reference scan to the reconstruction. For each 3D point in one~(either the reconstructed 3D points or the reference 3D points), its distance to the other is computed as the closest distance to all the 3D points in the other. The mean accuracy and completeness are recorded to evaluate the quality of a reconstruction. The evaluation code and the dataset can be downloaded on: \href{http://roboimagedata.compute.dtu.dk}{http://roboimagedata.compute.dtu.dk}

All experiments are conducted in a laptop with Intel 2.5GHz CPU and 8GB memory.

\begin{figure}[tb]
	\centering
	\includegraphics[width=0.5\textwidth]{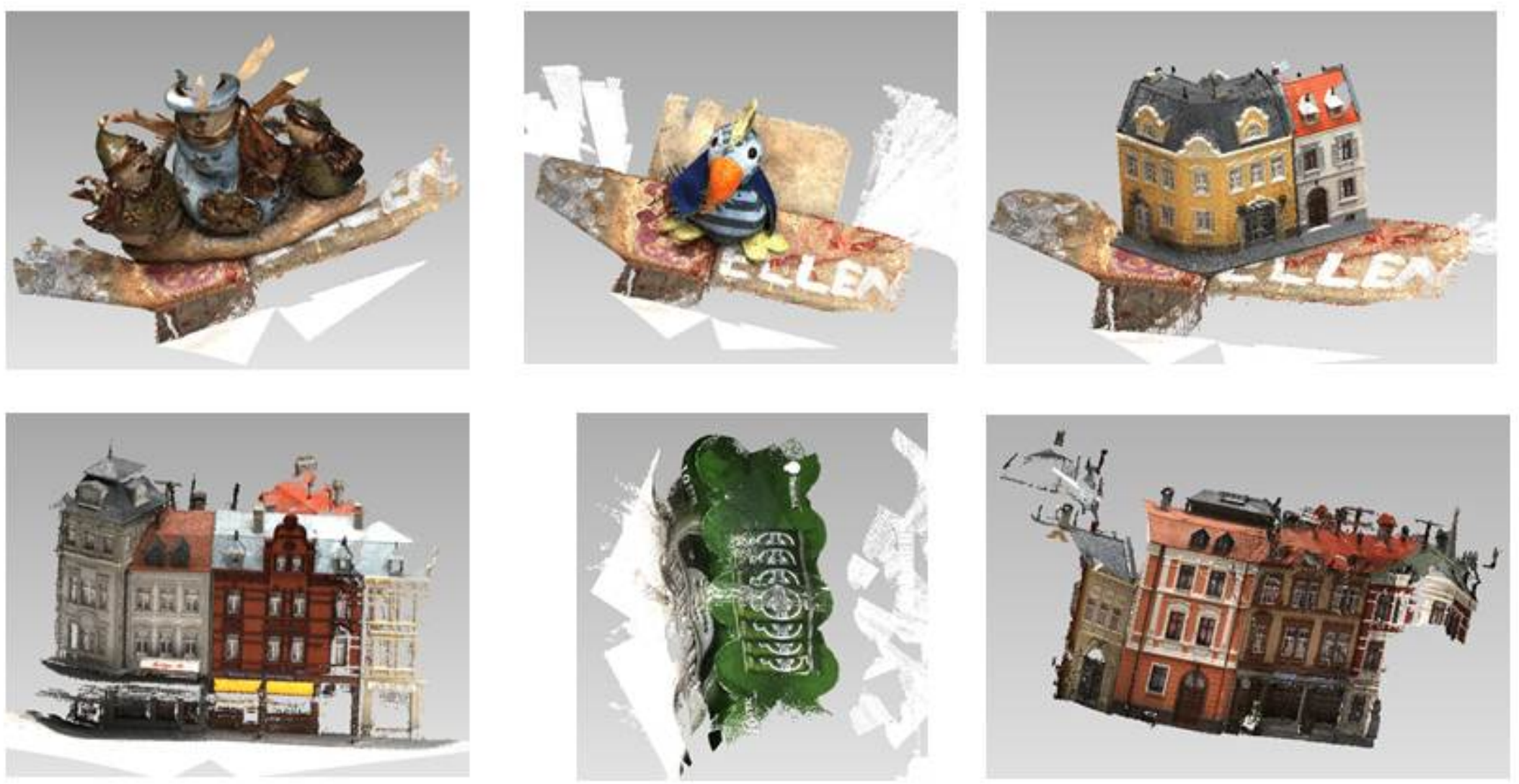}
	\caption{Some example scenes that have small performance difference for the evaluated methods.  \label{fig:smallvar_examples}}
\end{figure}

\subsection{Results and Analysis}

Among the 58 tested scenes, there are 3 scenes for which at least one method fails to obtain the reconstruction result due to the poor quality of point matching. For the remaining successful scenes, we further divide them into two groups. For one group, it contains scenes that all the evaluated methods perform similarly, i.e., both the variance of their reconstruction accuracy and the variance of their completeness are smaller than a threshold, which we set to be 0.05. For the other group, it contains those scenes that all the evaluated methods have a large variance of their performance, i.e., at least one method performs significantly different from other ones. There are 6 scenes in this group. We will analysis the performance of the evaluated methods for these three groups of scenes respectively.

\begin{figure}[tb]
	\centering
    \subfloat[]{
		\includegraphics[width=0.235\textwidth]{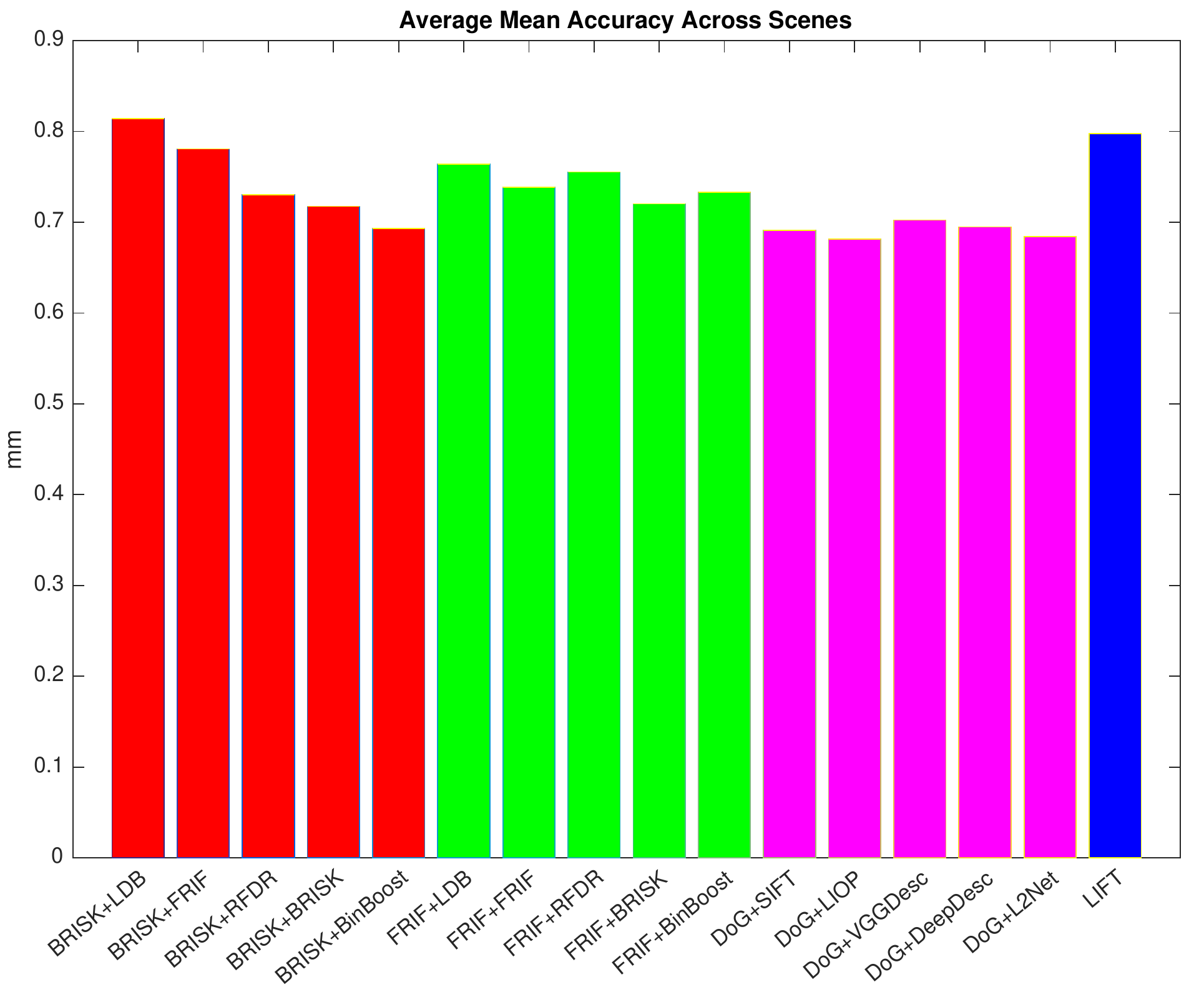}}
	\subfloat[]{
		\includegraphics[width=0.235\textwidth]{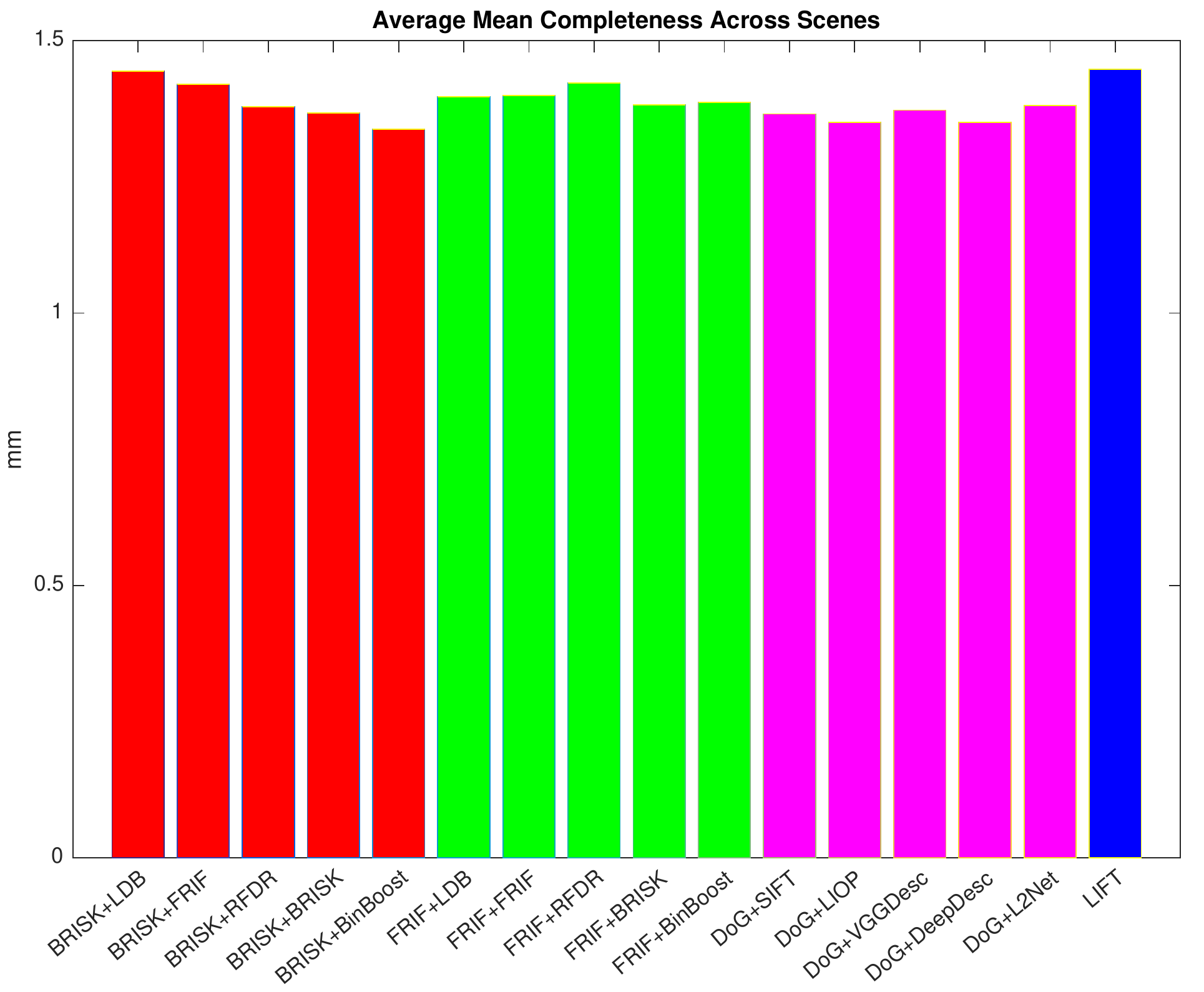}}
	\caption{The average reconstruction (a)~accuracy and (b)~completeness over all the scenes that have small performance variance for the evaluated methods. See text for details. \label{fig:smallvar_result}}
\end{figure}

\begin{figure}[tb]
	\centering
    \subfloat[]{
		\includegraphics[width=0.23\textwidth]{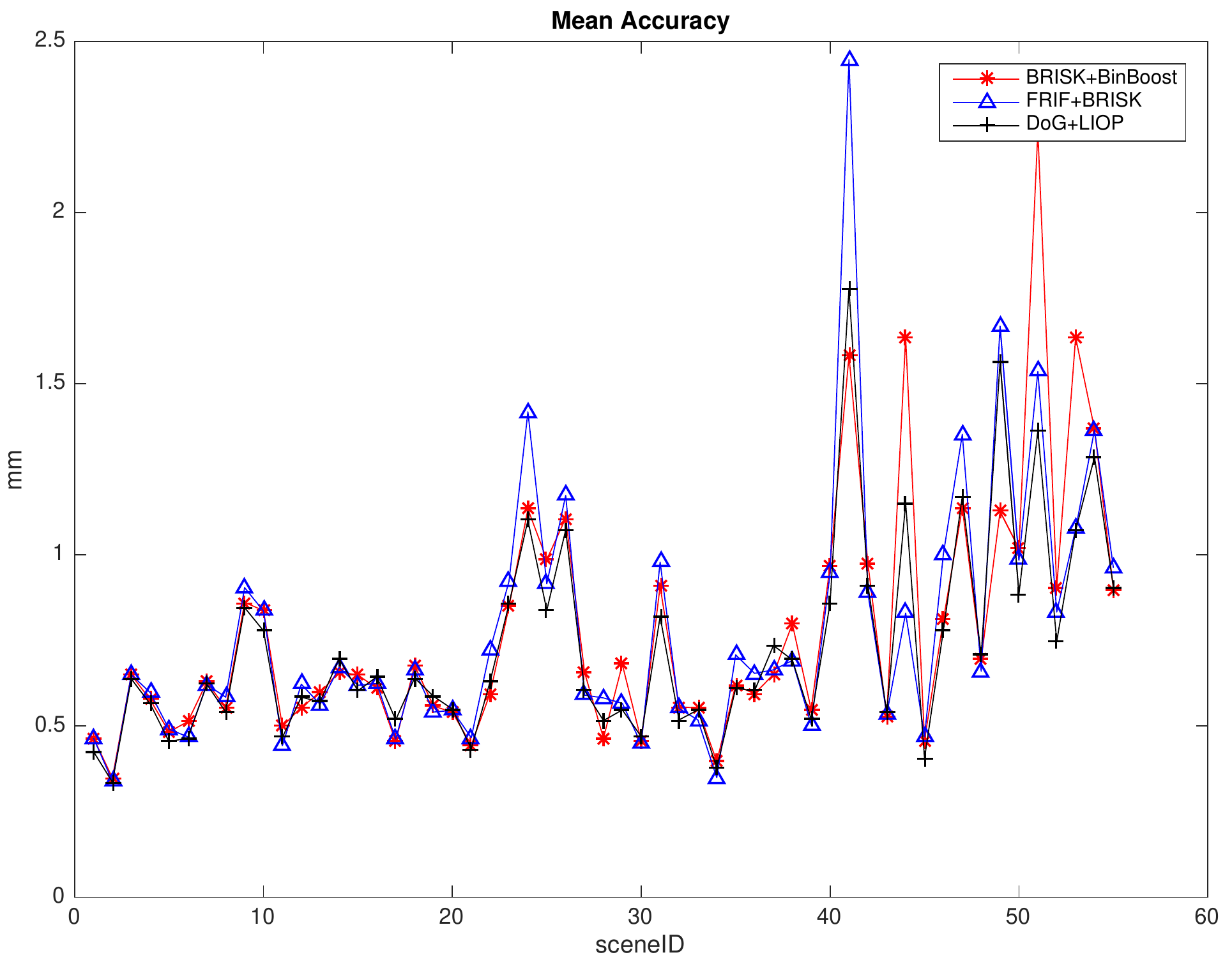}}
	\subfloat[]{
		\includegraphics[width=0.23\textwidth]{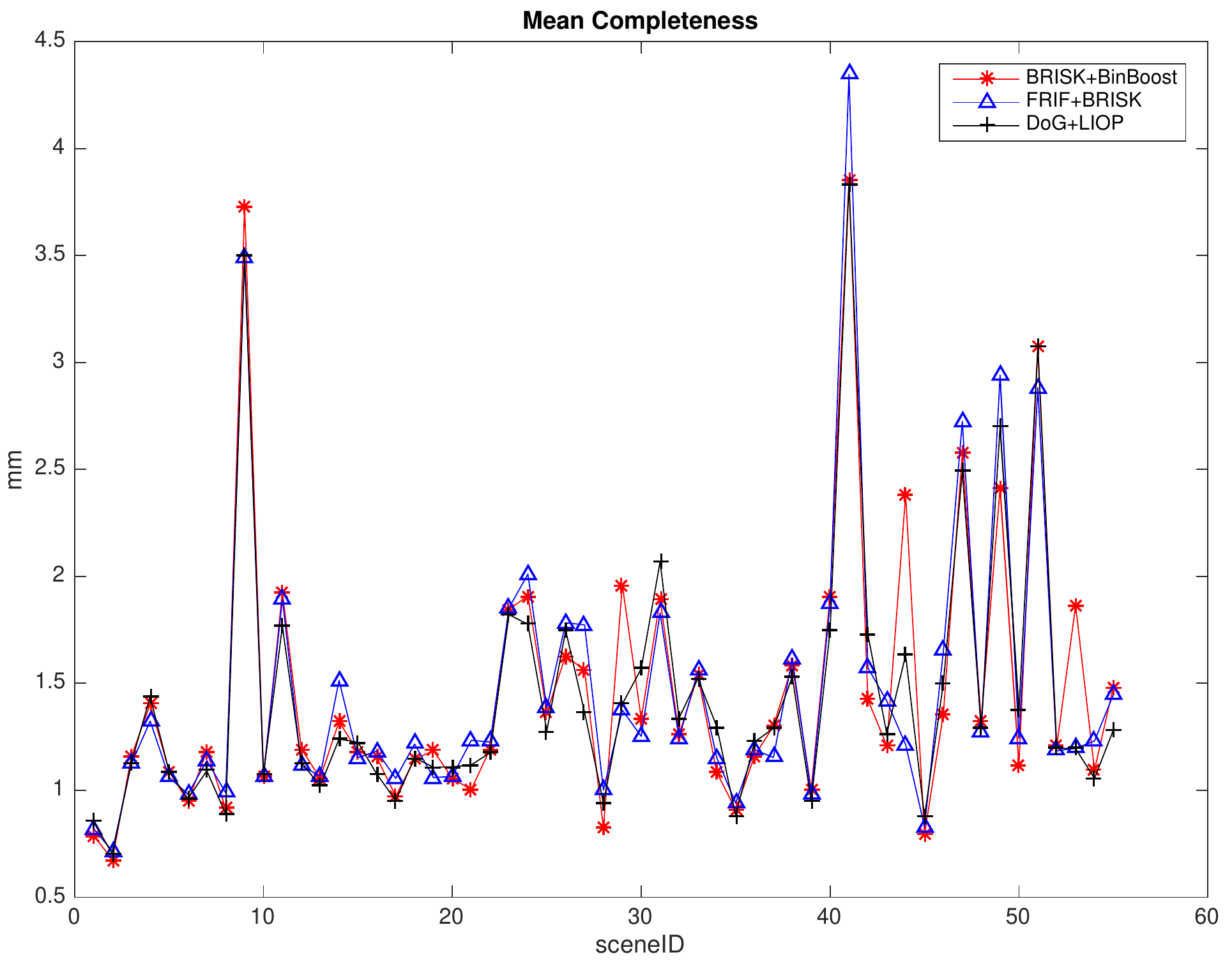}}
	\caption{The average reconstruction (a)~accuracy and (b)~completeness for the 55 different scenes that all the evaluated methods successfully obtained reconstruction results. To reduce cluster, we show results for each keypoint with the best descriptor combination. \label{fig:all_success_result}}
\end{figure}

\textbf{\emph{Scenes with small performance variance}}. In this case, it corresponds to the easiest scenes for 3D reconstruction. Some examples of these scenes are shown in Fig.~\ref{fig:smallvar_examples}. These scenes all contain rich textures and are easy for feature point matching. The average mean accuracy of different methods across all scenes of this kind~(i.e., with small performance variance) is shown in Fig.~\ref{fig:smallvar_result}(a), while the average mean completeness is shown in Fig.~\ref{fig:smallvar_result}(b). Among the binary features, the combination of BRISK keypoint with BinBoost descriptor performs the best, whose performance is comparable or even better than some float features. For all the tested combinations, BRISK keypoint with BinBoost descriptor and DoG keypoint with LIOP descriptor perform similar, both of which are with the top performance. In general, DoG with float descriptors lead to a better reconstruction accuracy than using binary features, except for the best combination of BRISK + BinBoost. An interesting observation is that the entire feature learning solution, LIFT, does not perform as well as other float features. In fact, it performs the worst among all the evaluated features, including the binary ones. Obviously, using LIFT leads to larger reconstruction error both in terms of accuracy and completeness. Such an inferior performance of LIFT indicates that there might be larger localization error between corresponding LIFT keypoints since it indeed produces comparable or more matching points than SIFT in our experiments. Except for LIFT, LIOP produces slightly better results than other float type descriptors and the remaining ones perform similarly. Among all the binary descriptors, LDB is not as good as others no matter which keypoint is used. Meanwhile, when using FRIF keypoint, the results of different binary descriptors are more flat than using BRISK keypoint. This means that FRIF keypoint is less sensitive to descriptors. For BRISK keypoint, it has to be careful when choosing the combined descriptor so as to achieve good performance. From Fig.~\ref{fig:smallvar_result}, we can conclude that it is not necessary to learn sophisticated descriptors for easy scenes. In this case, using binary features is good enough to obtain satisfactory reconstruction accuracy as using float features.  Taking the best descriptor for each keypoint, we show mean accuracy and completeness of all 55 successful scenes~(all the evaluated methods successfully obtained reconstruction results) in Fig.~\ref{fig:all_success_result}. Note that we do not include the results of LIFT as it performs the worst according to the average results.

\begin{figure*}
\centering
	\subfloat[]{
	\begin{minipage}[c]{0.12\textwidth}
		\centering
		\includegraphics[width=1.0\textwidth]{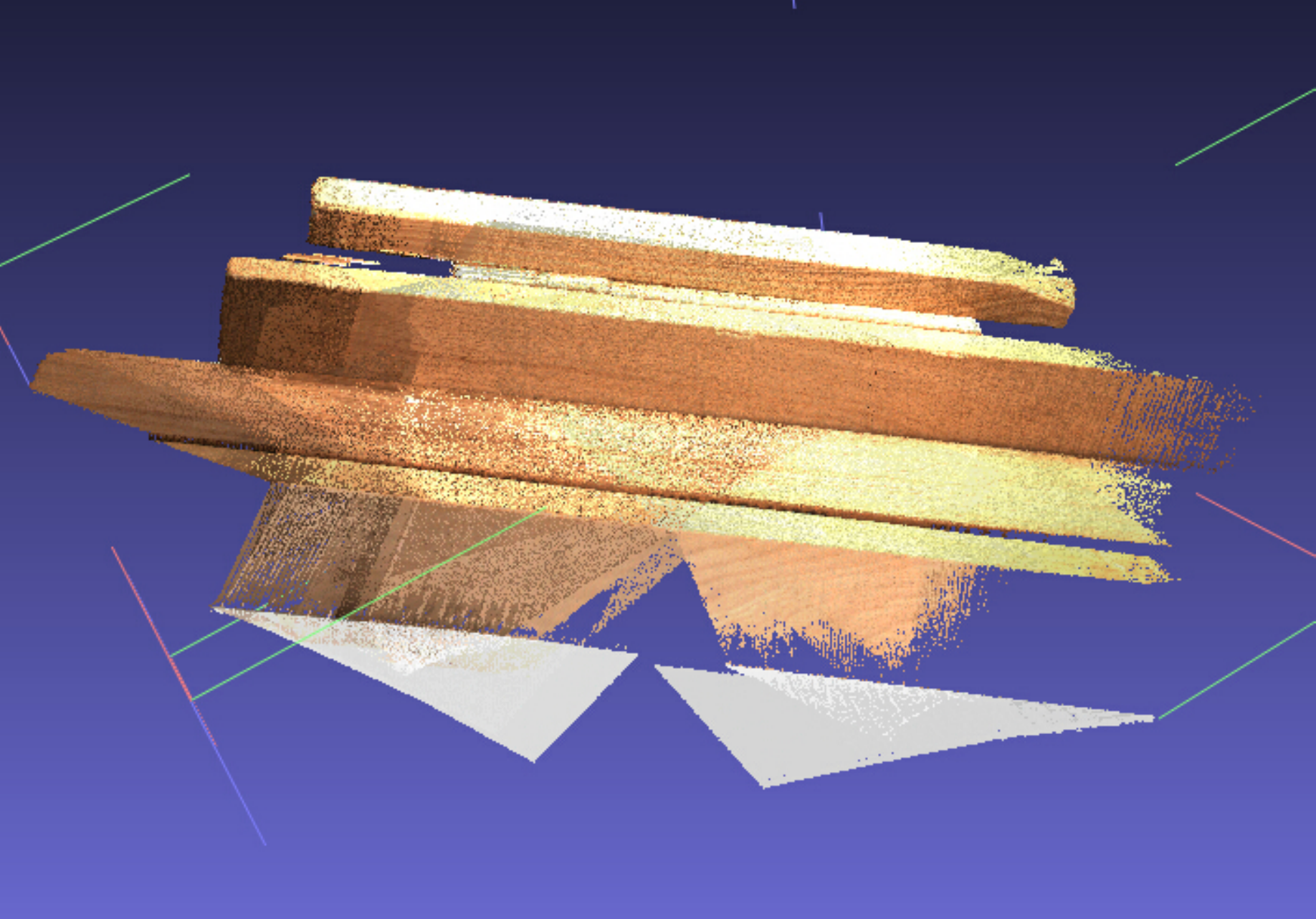}
	\end{minipage}
	\begin{minipage}[c]{0.78\textwidth}
		\centering
		\includegraphics[width=0.28\textwidth]{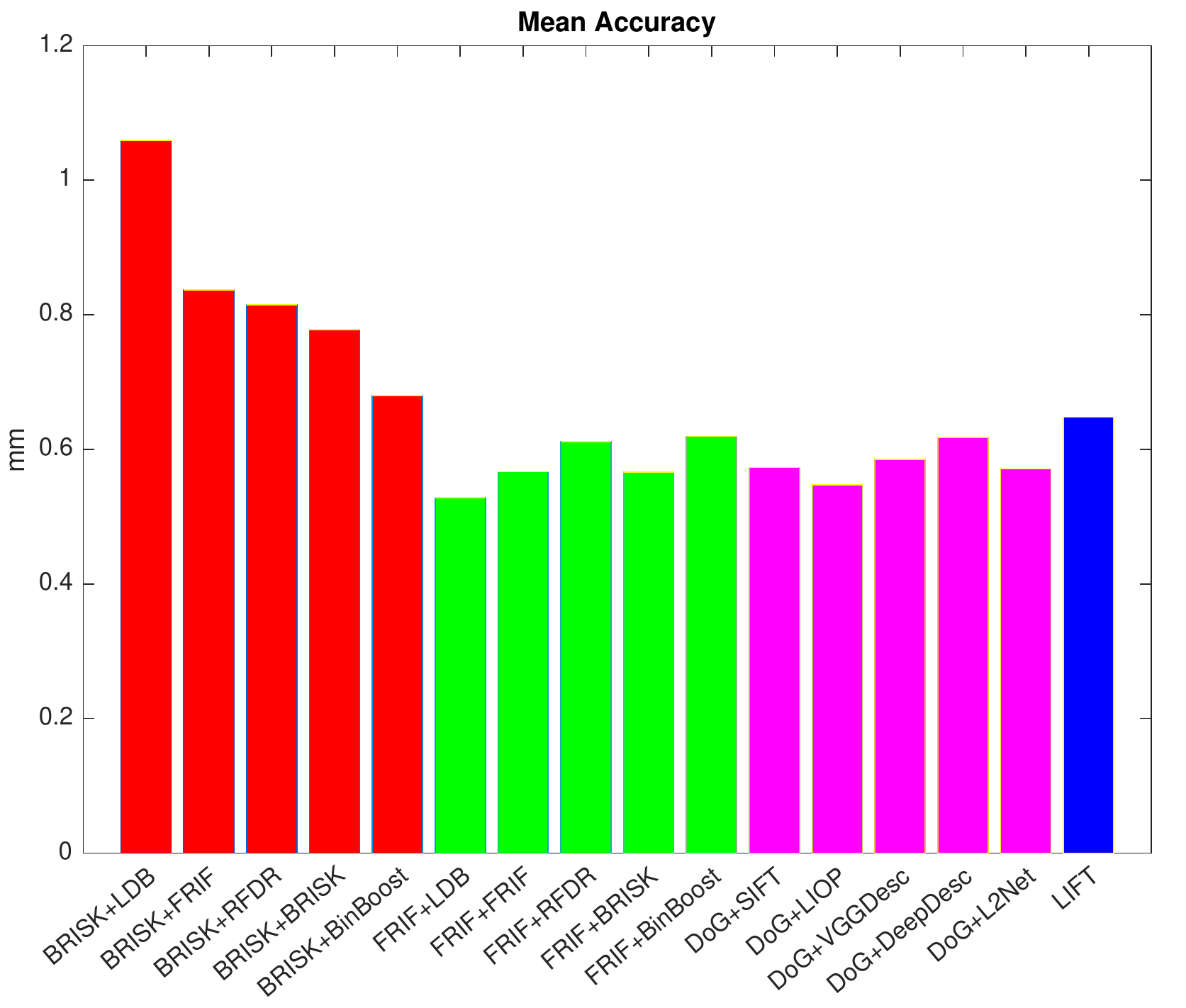}
		\includegraphics[width=0.28\textwidth]{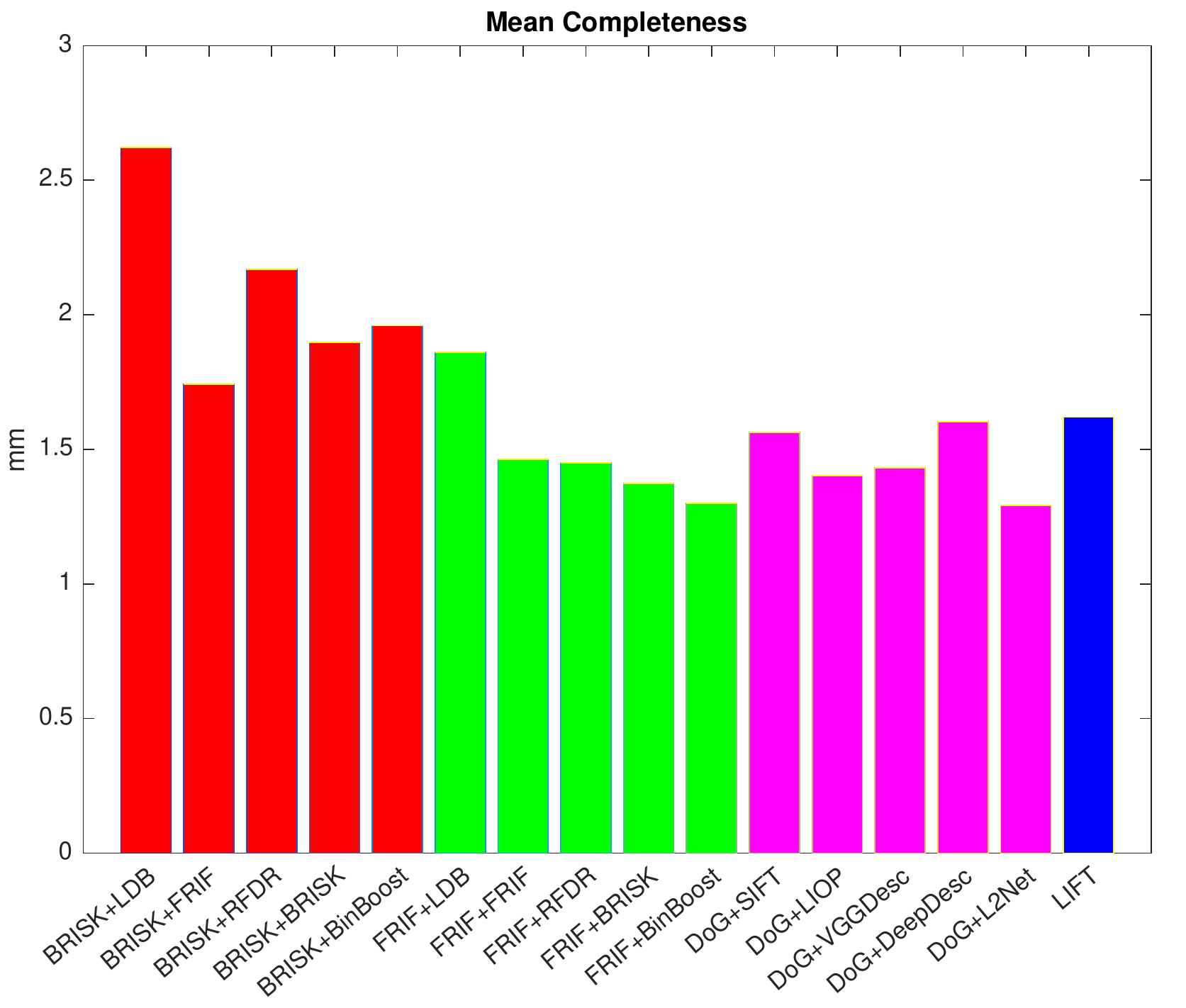}
		\includegraphics[width=0.28\textwidth]{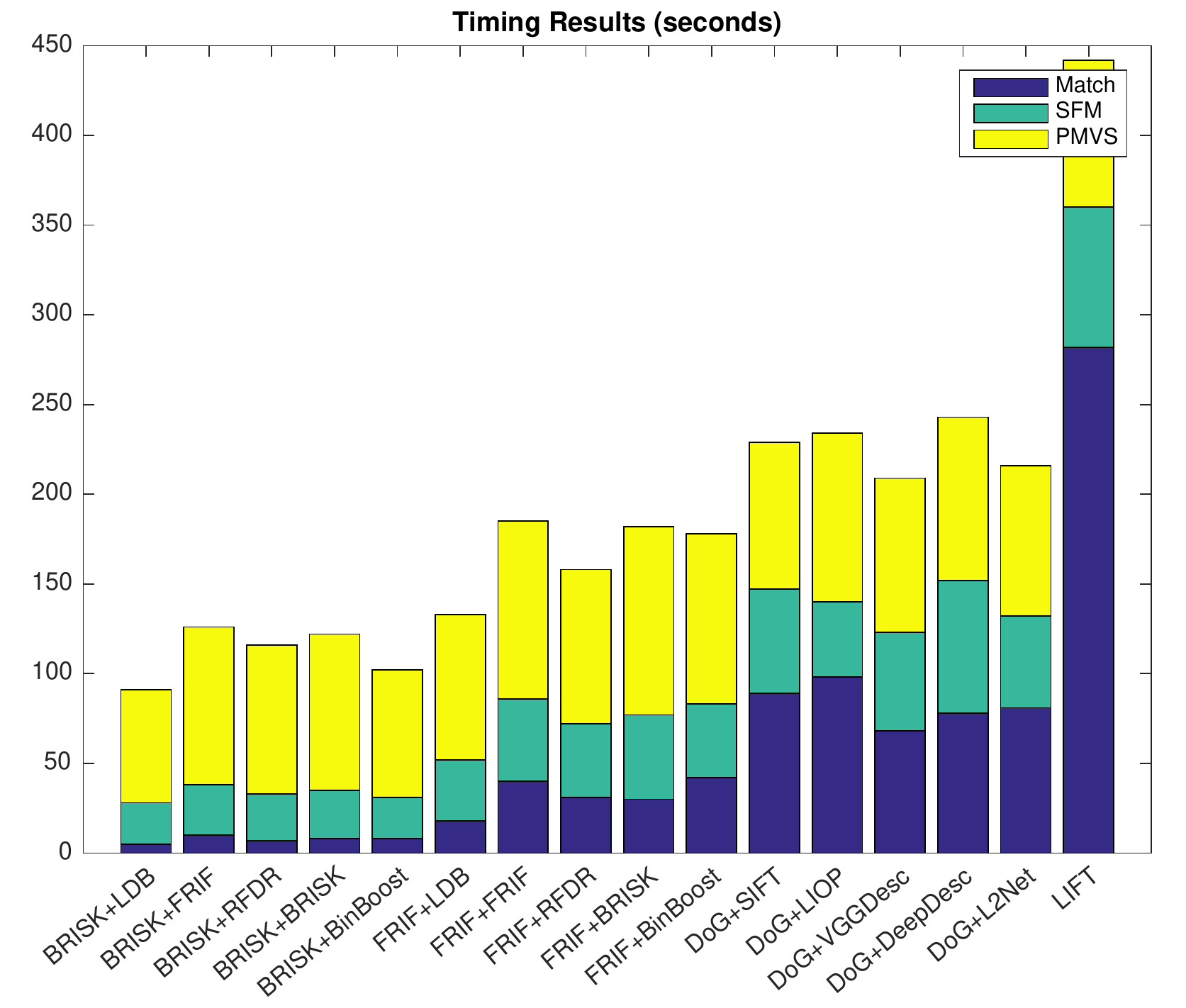}
	\end{minipage}}
		
	\subfloat[]{
	\begin{minipage}[c]{0.12\textwidth}
		\centering
		\includegraphics[width=1.0\textwidth]{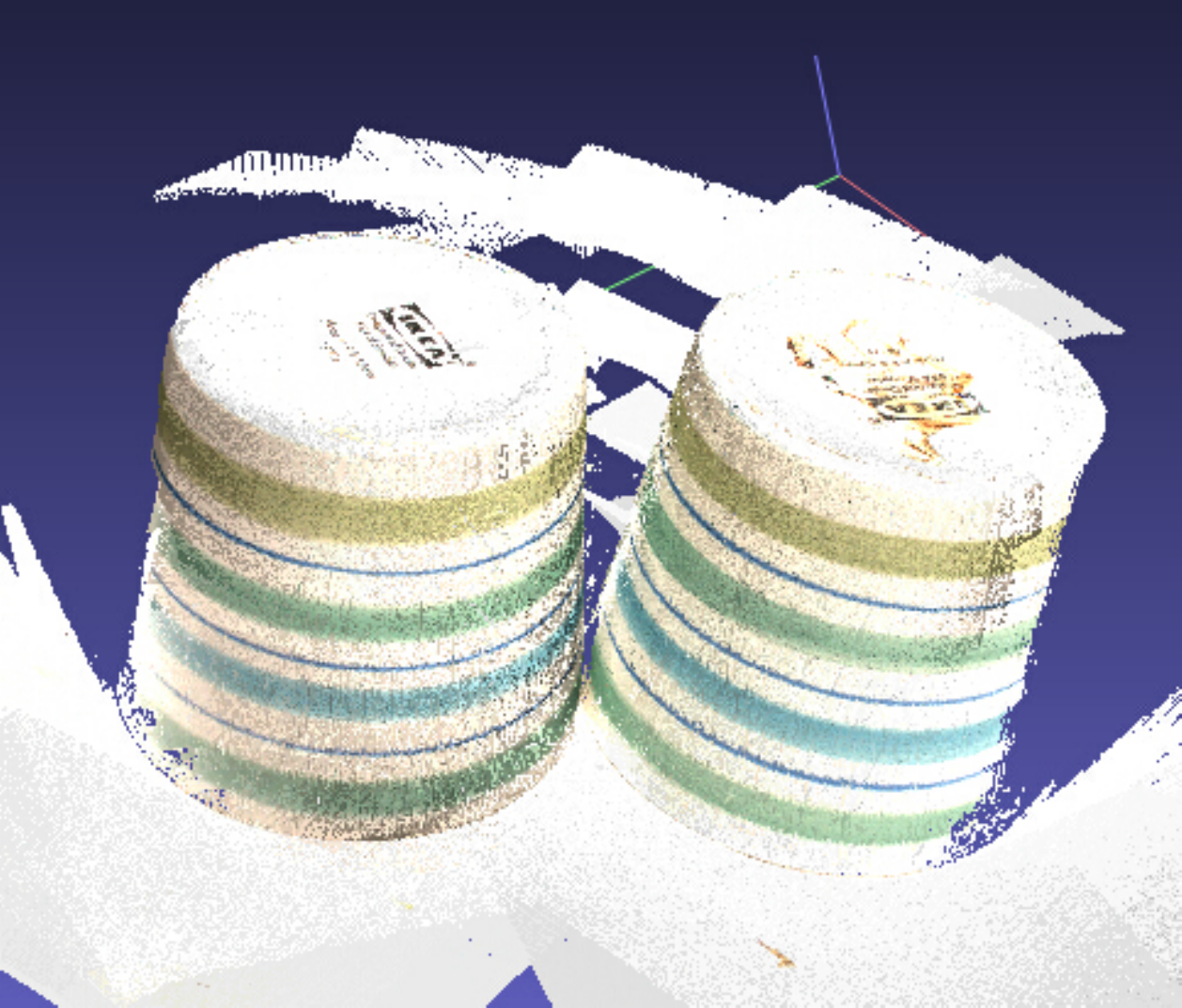}
	\end{minipage}
	\begin{minipage}[c]{0.78\textwidth}
		\centering
		\includegraphics[width=0.28\textwidth]{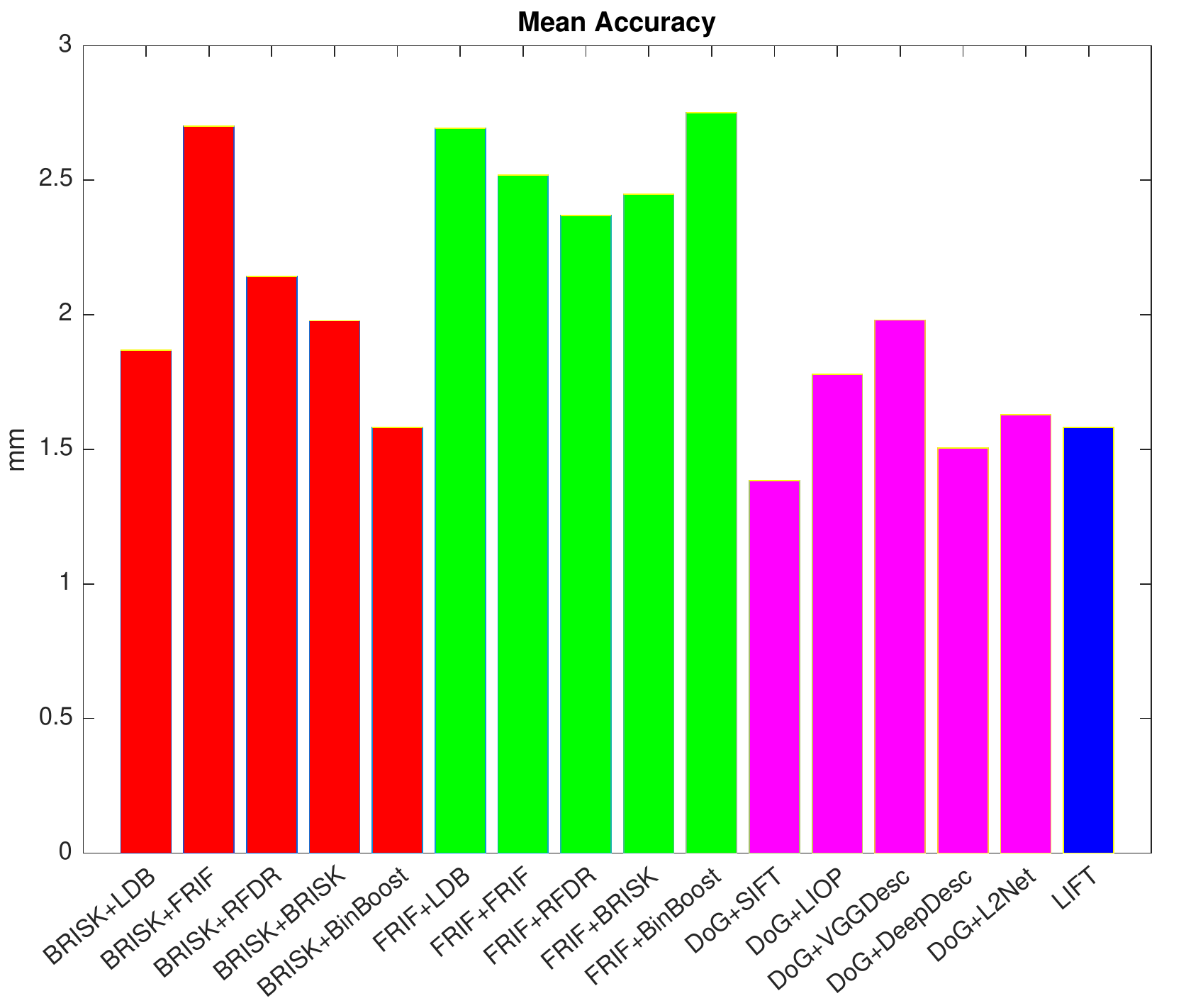}
		\includegraphics[width=0.28\textwidth]{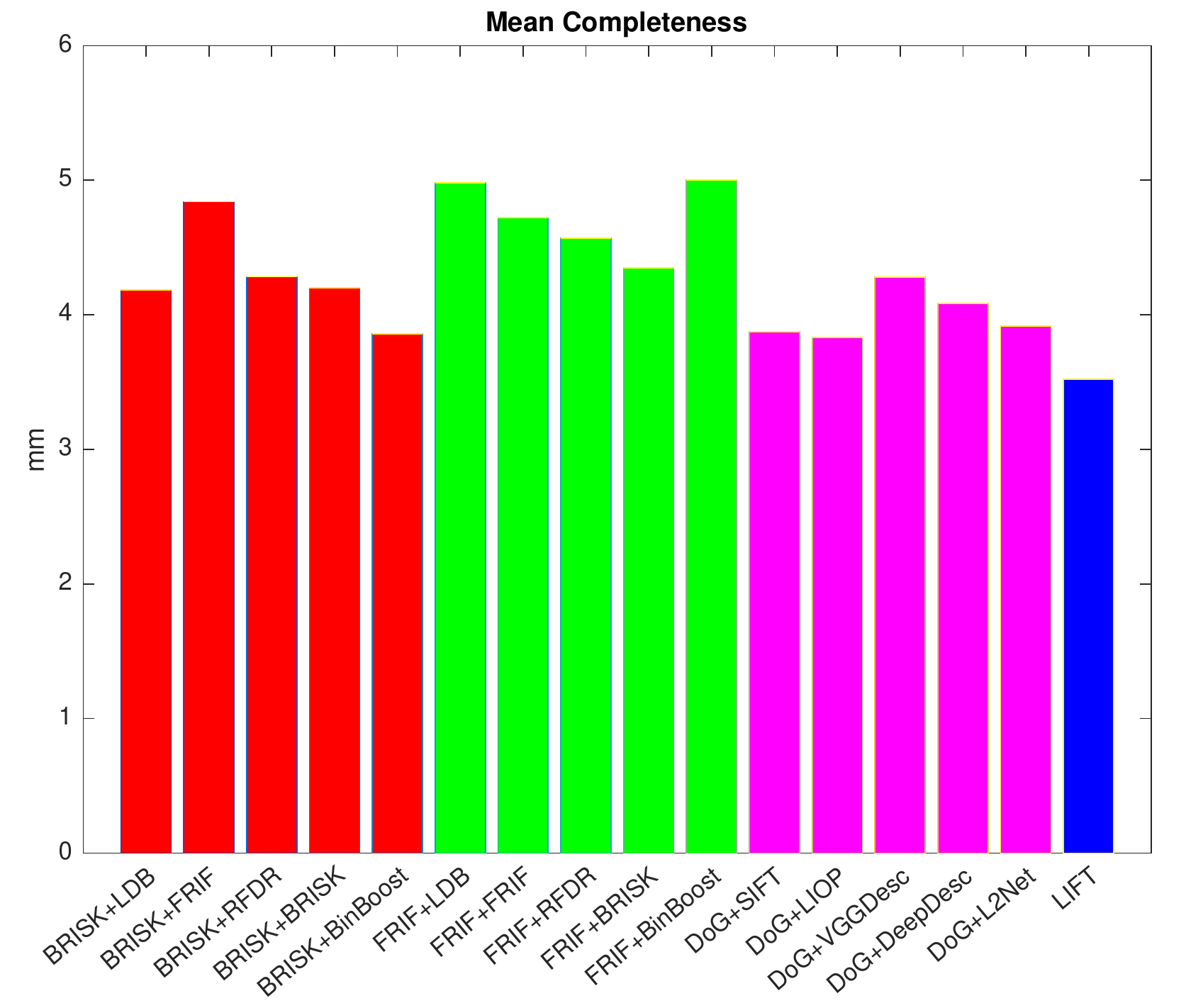}
		\includegraphics[width=0.28\textwidth]{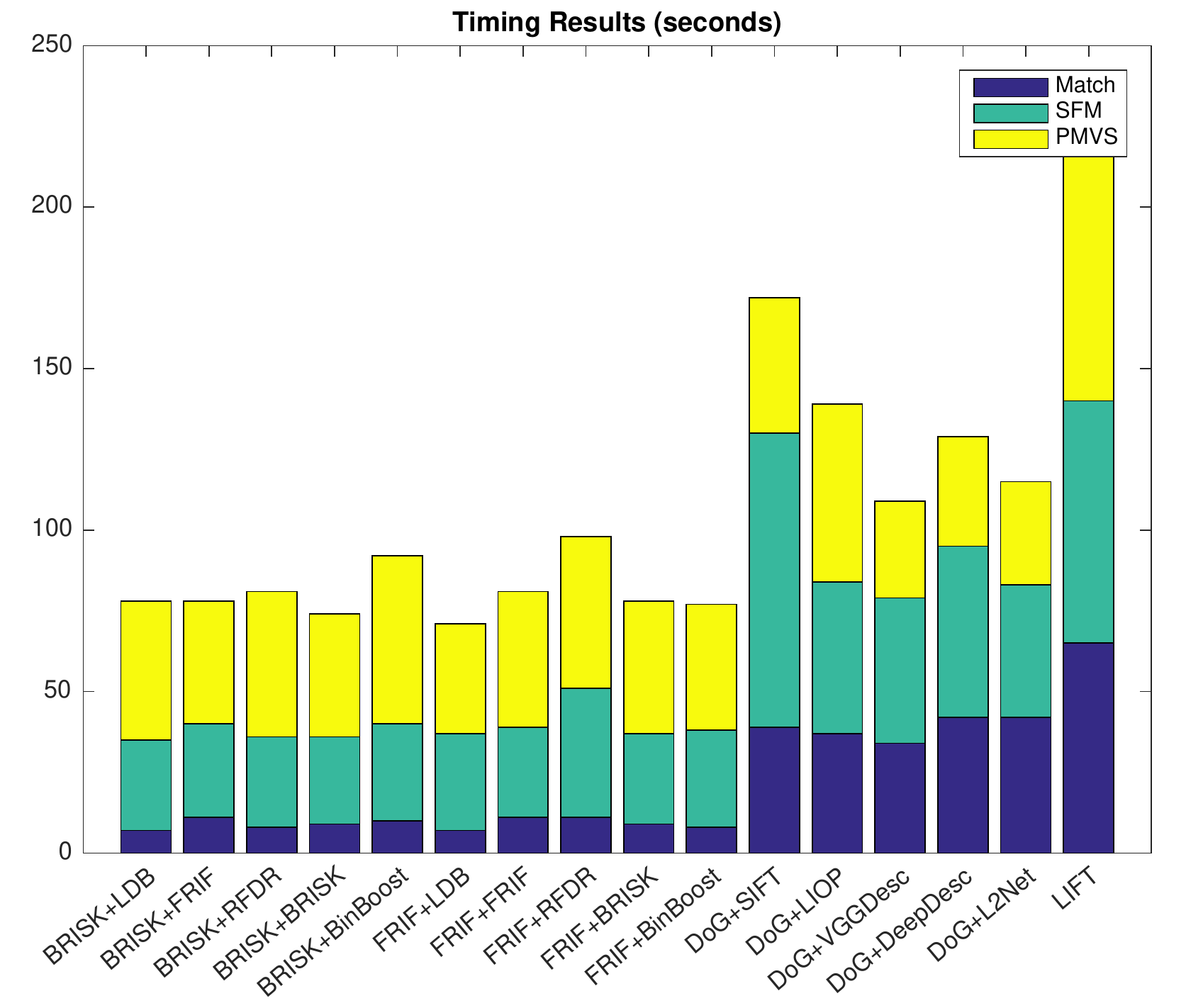}
	\end{minipage}}
		
	\subfloat[]{
	\begin{minipage}[c]{0.12\textwidth}
		\centering
		\includegraphics[width=1.0\textwidth]{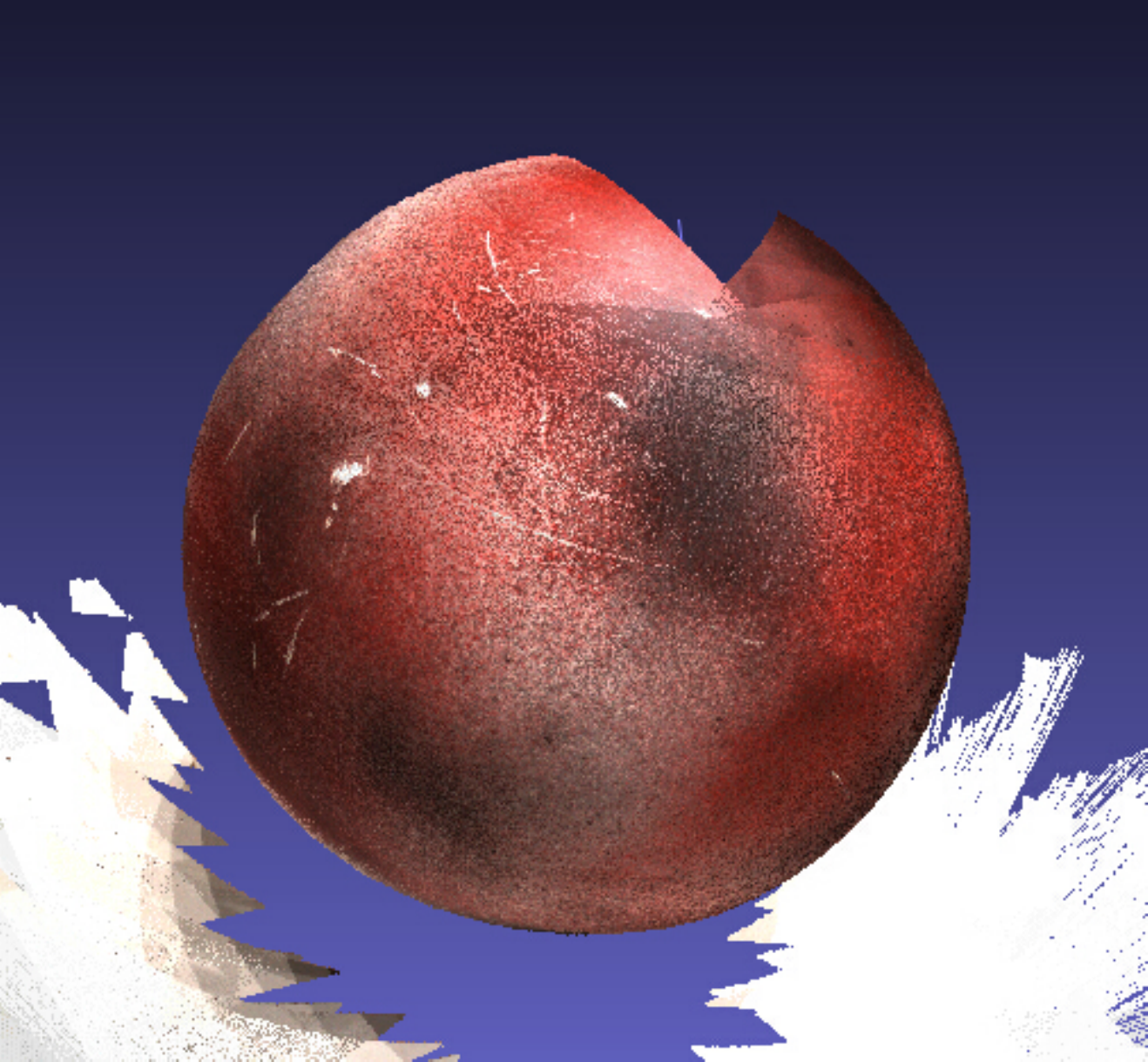}
	\end{minipage}
	\begin{minipage}[c]{0.78\textwidth}
		\centering
		\includegraphics[width=0.28\textwidth]{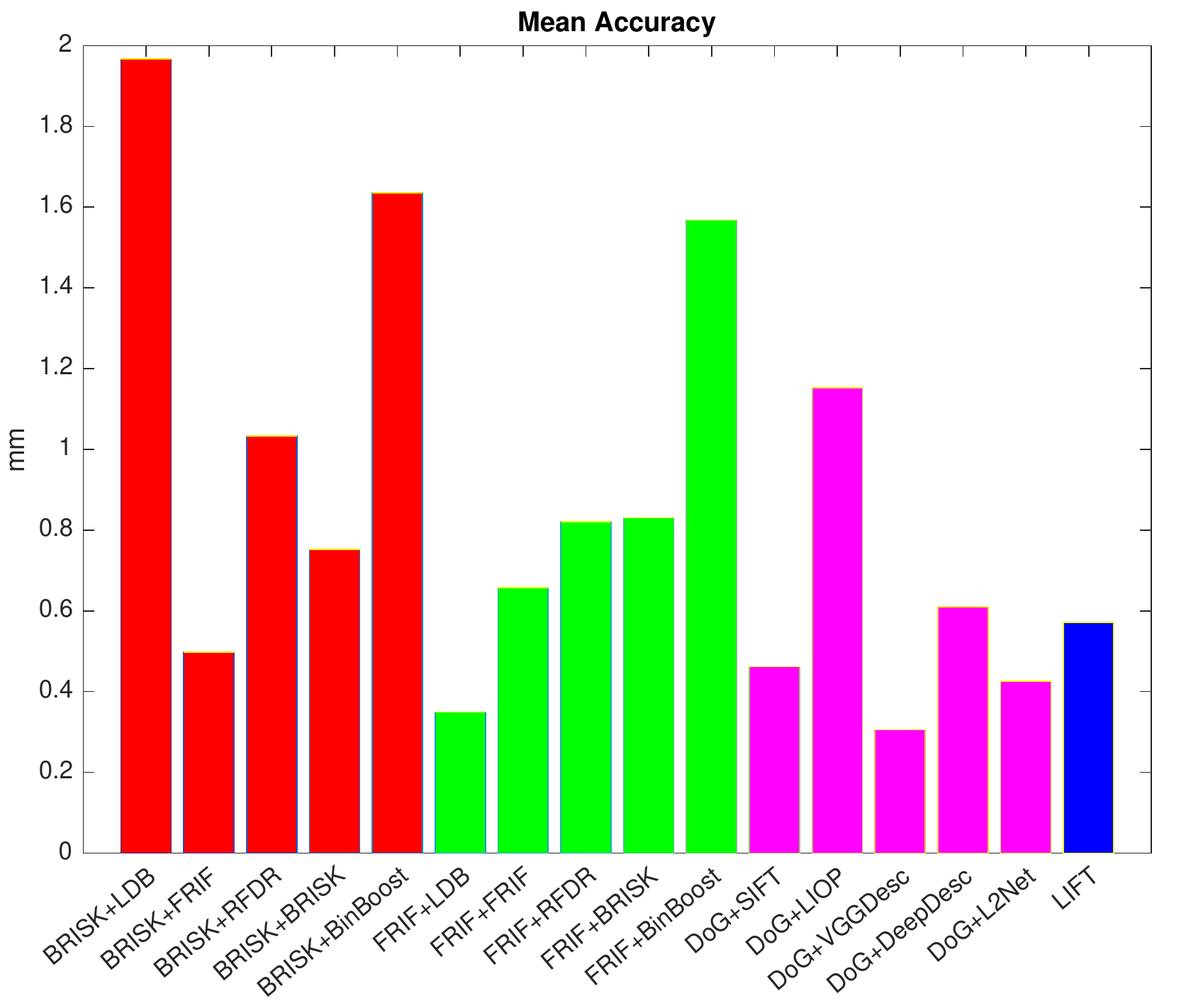}
		\includegraphics[width=0.28\textwidth]{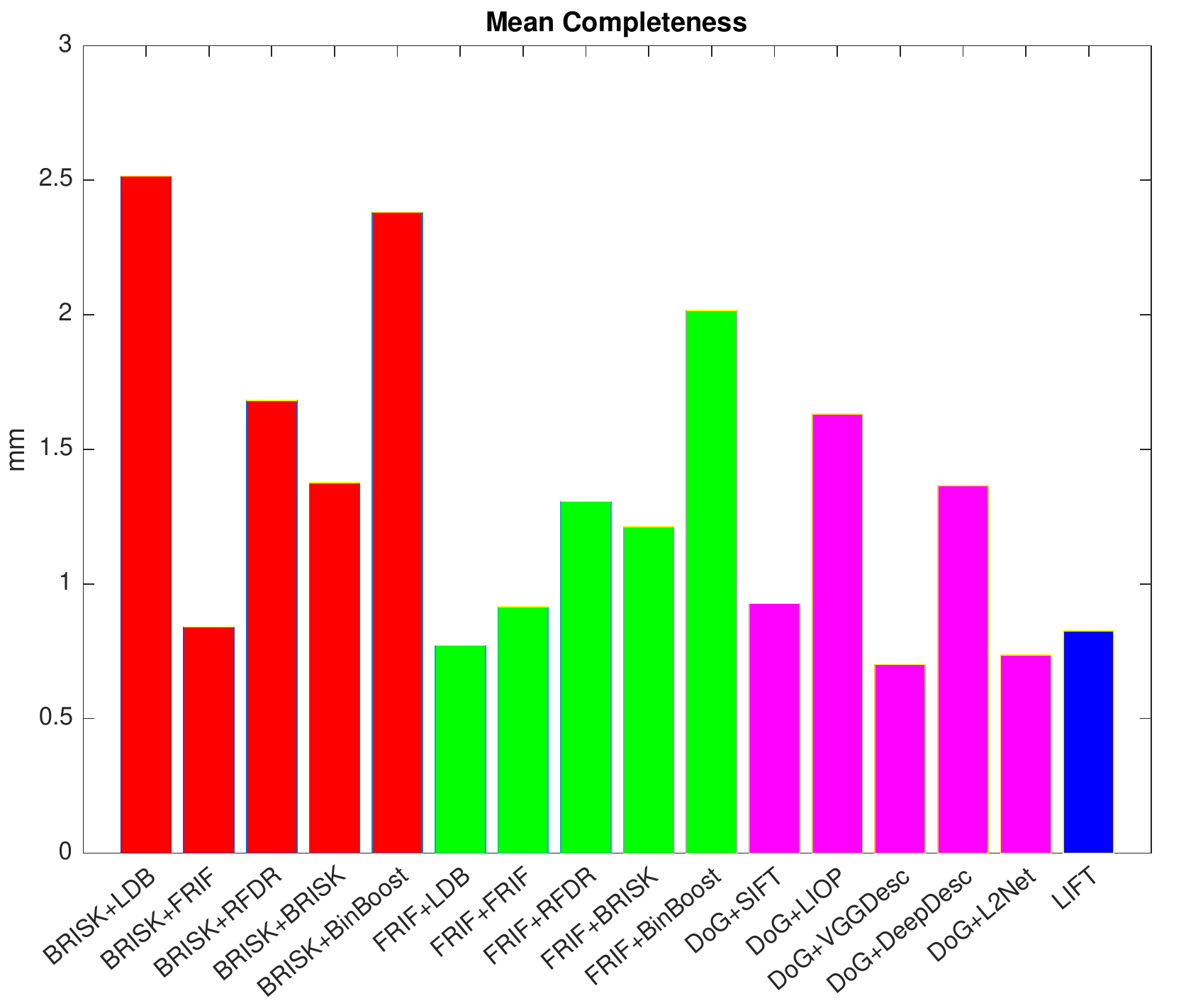}
		\includegraphics[width=0.28\textwidth]{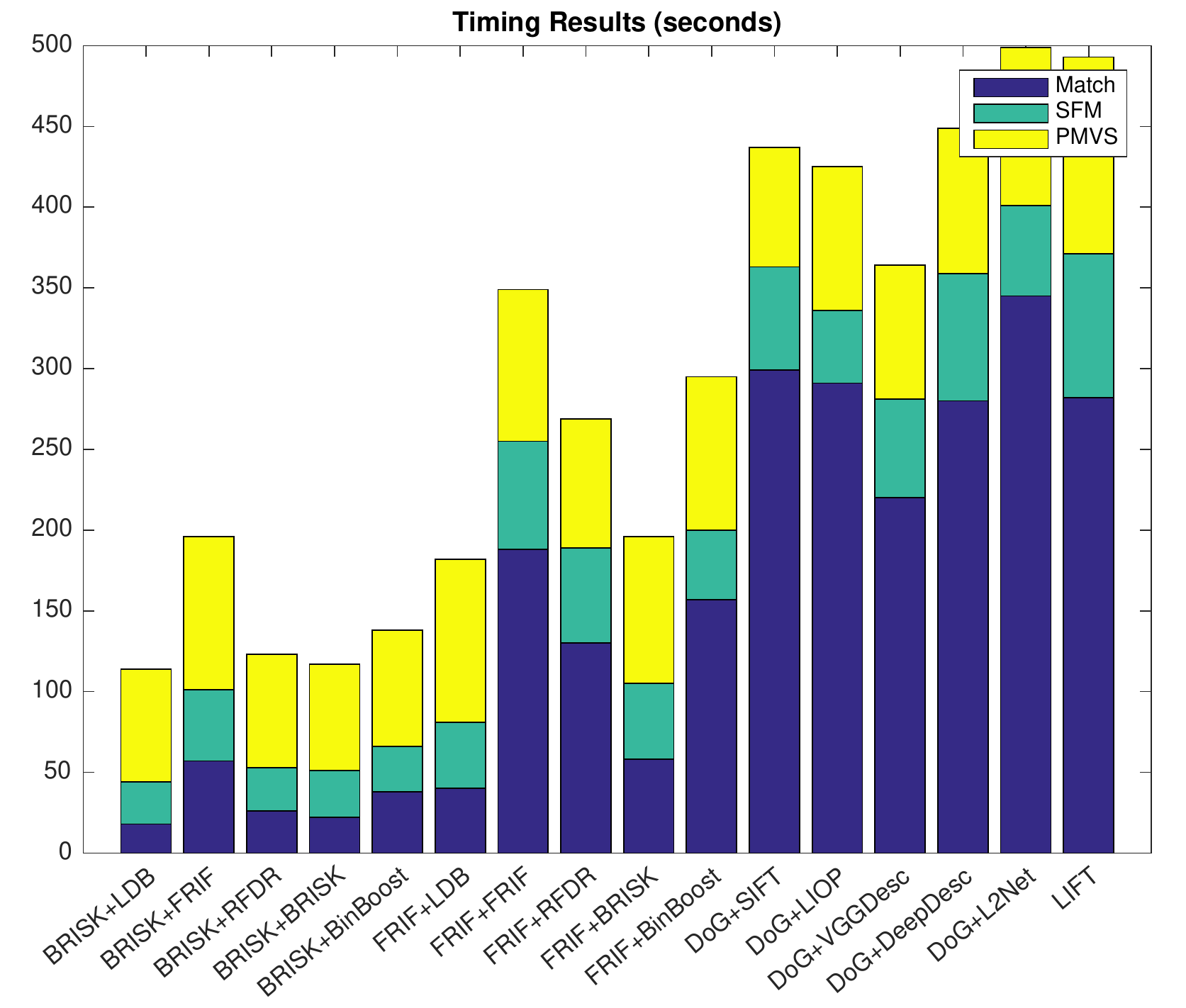}
	\end{minipage}}

    \subfloat[]{
	\begin{minipage}[c]{0.12\textwidth}
		\centering
		\includegraphics[width=1.0\textwidth]{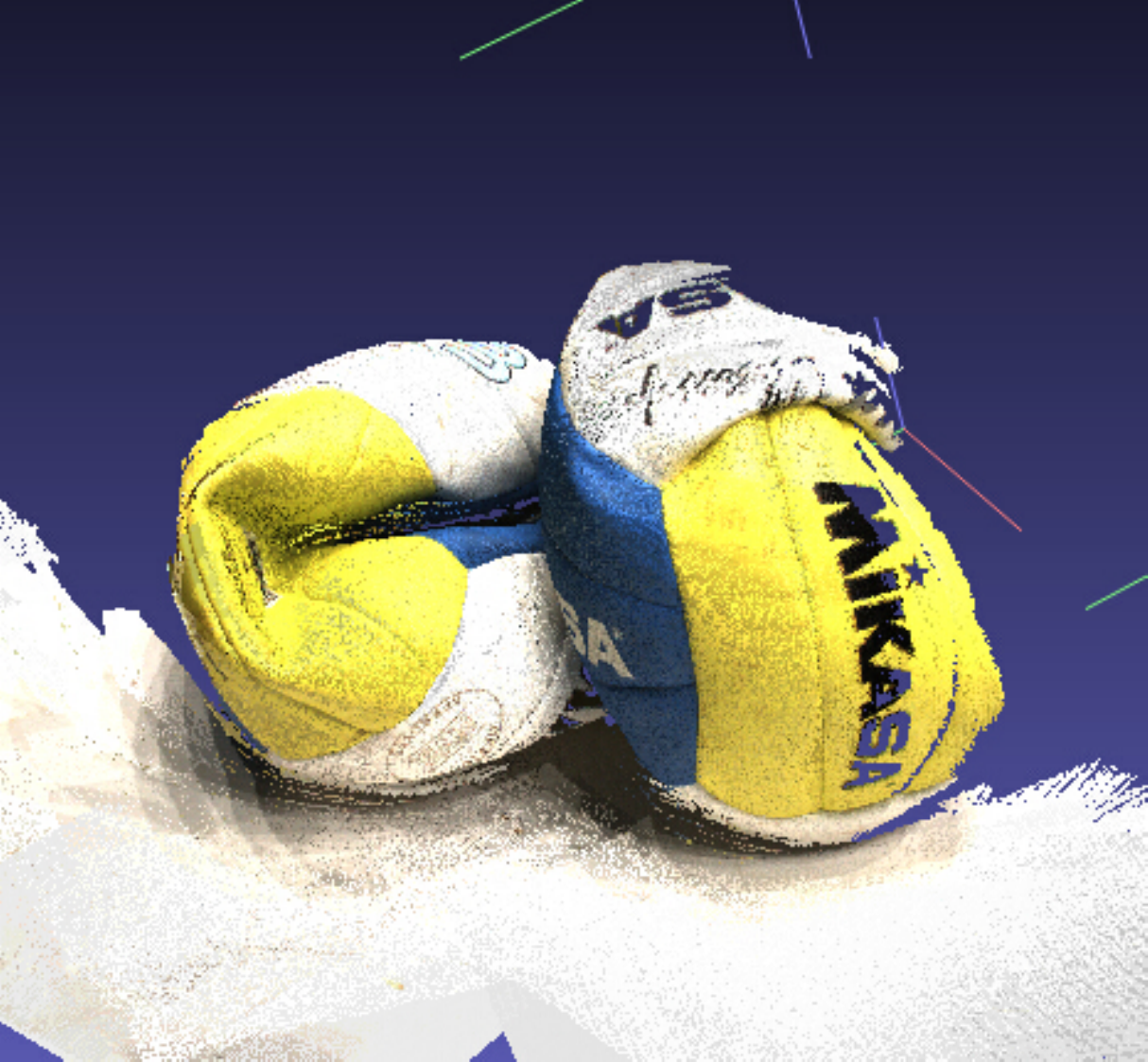}
	\end{minipage}
	\begin{minipage}[c]{0.78\textwidth}
		\centering
		\includegraphics[width=0.28\textwidth]{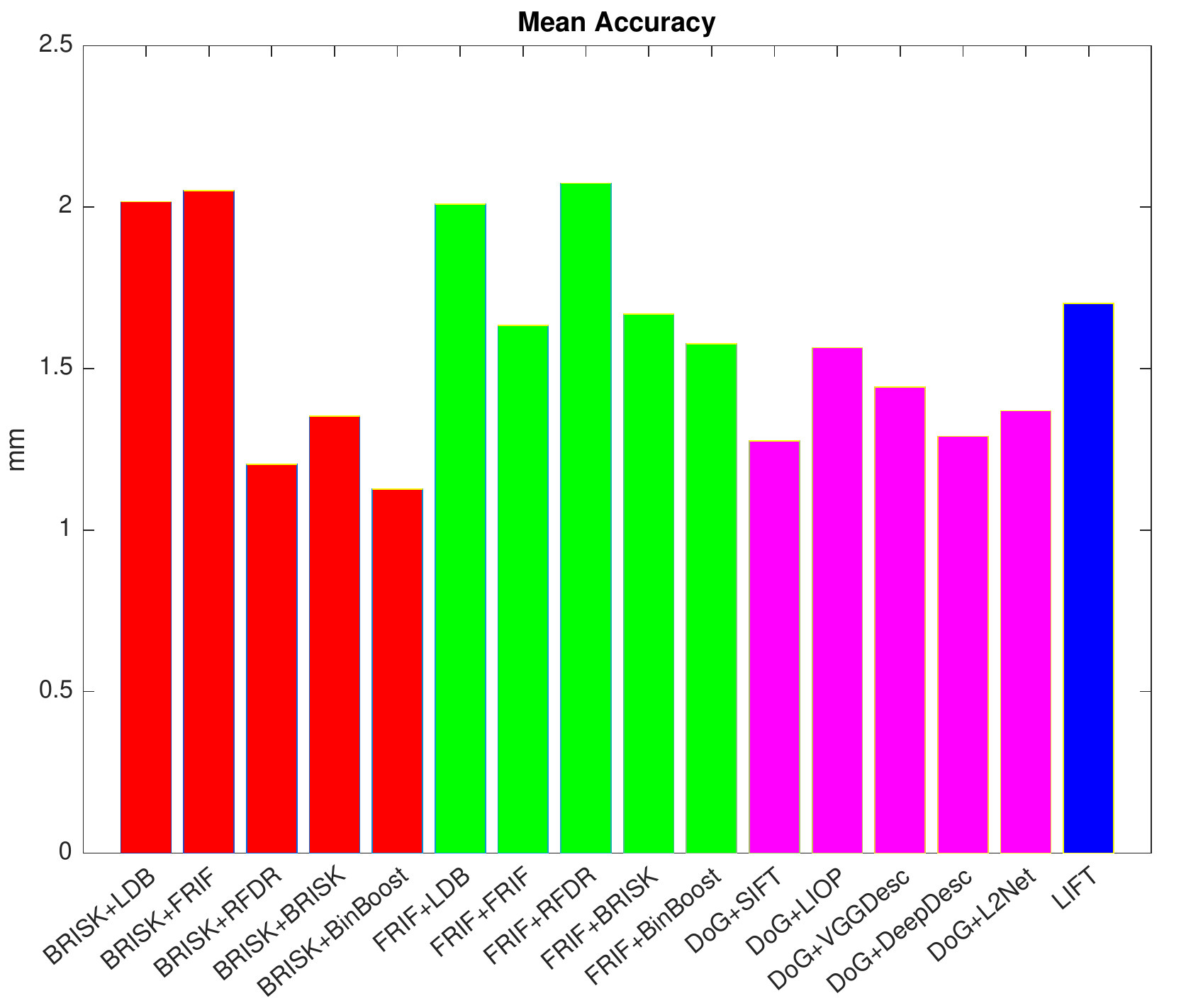}
		\includegraphics[width=0.28\textwidth]{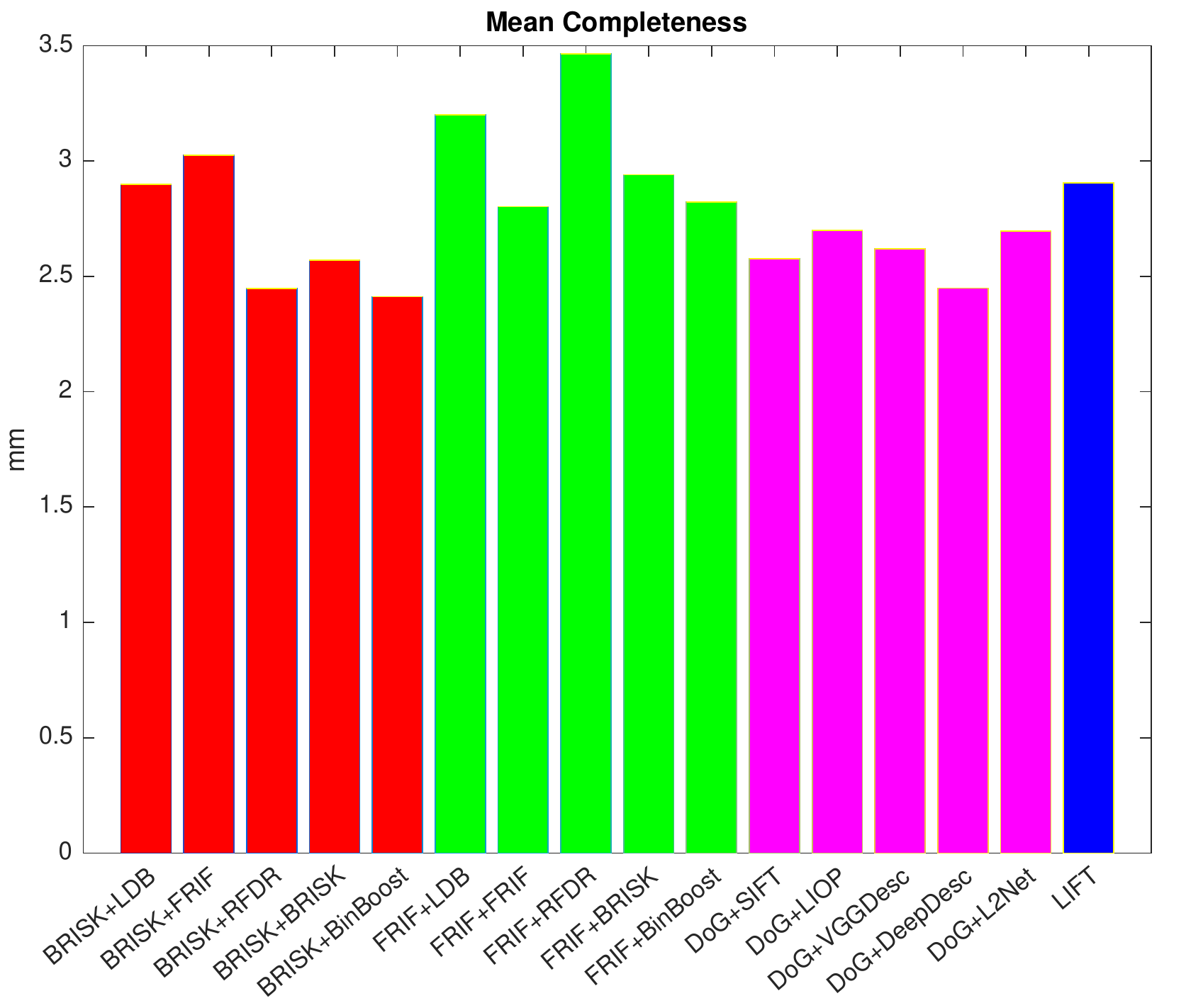}
		\includegraphics[width=0.28\textwidth]{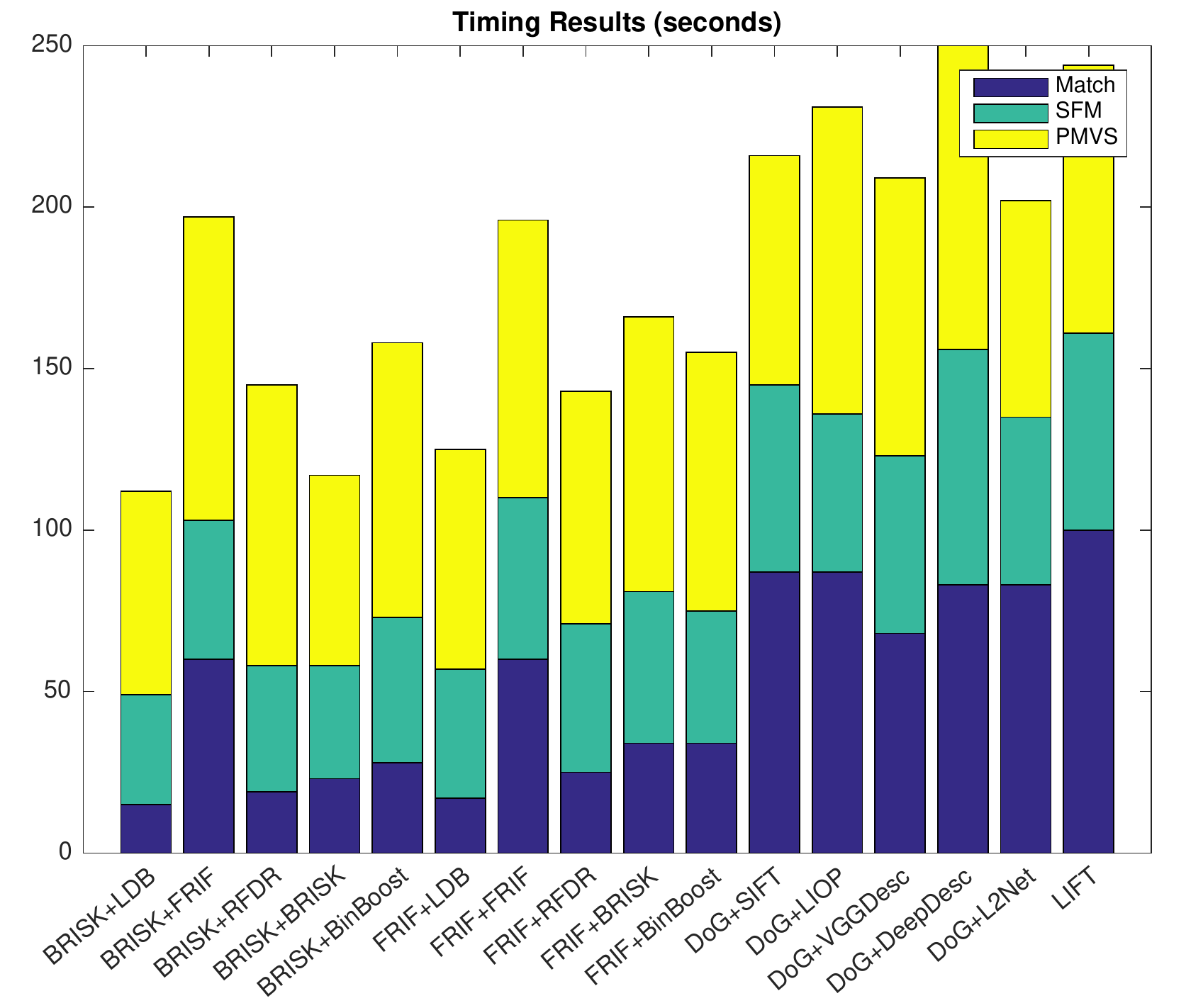}
	\end{minipage}}

    \subfloat[]{
	\begin{minipage}[c]{0.12\textwidth}
		\centering
		\includegraphics[width=1.0\textwidth]{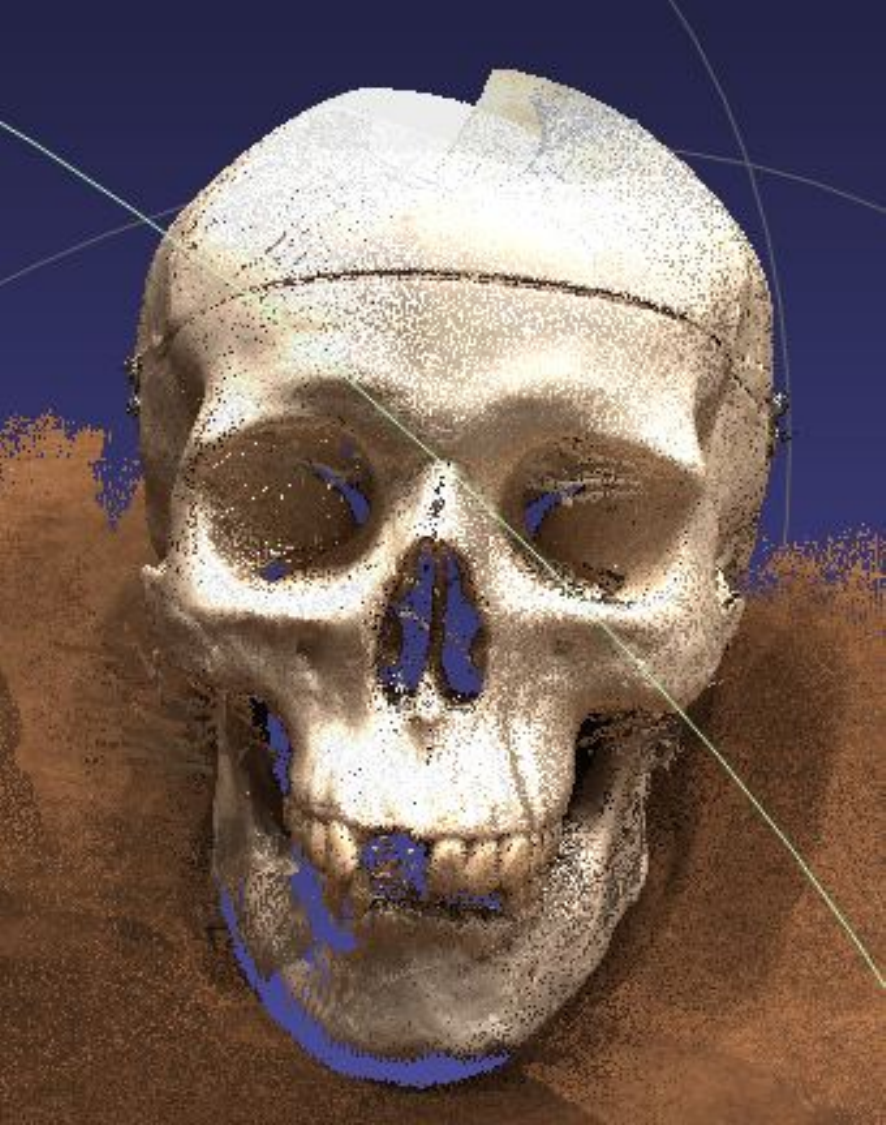}
	\end{minipage}
	\begin{minipage}[c]{0.78\textwidth}
		\centering
		\includegraphics[width=0.28\textwidth]{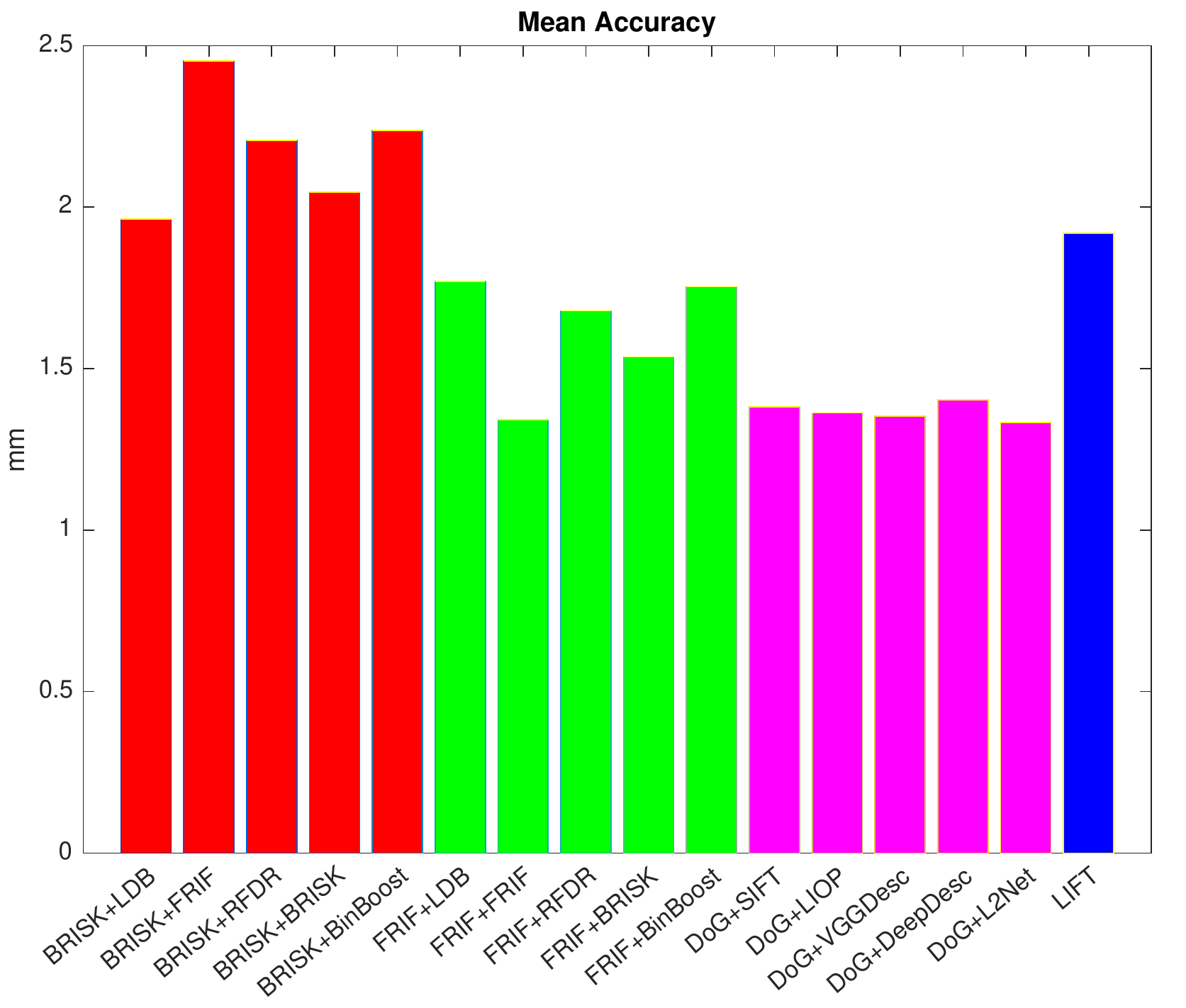}
		\includegraphics[width=0.28\textwidth]{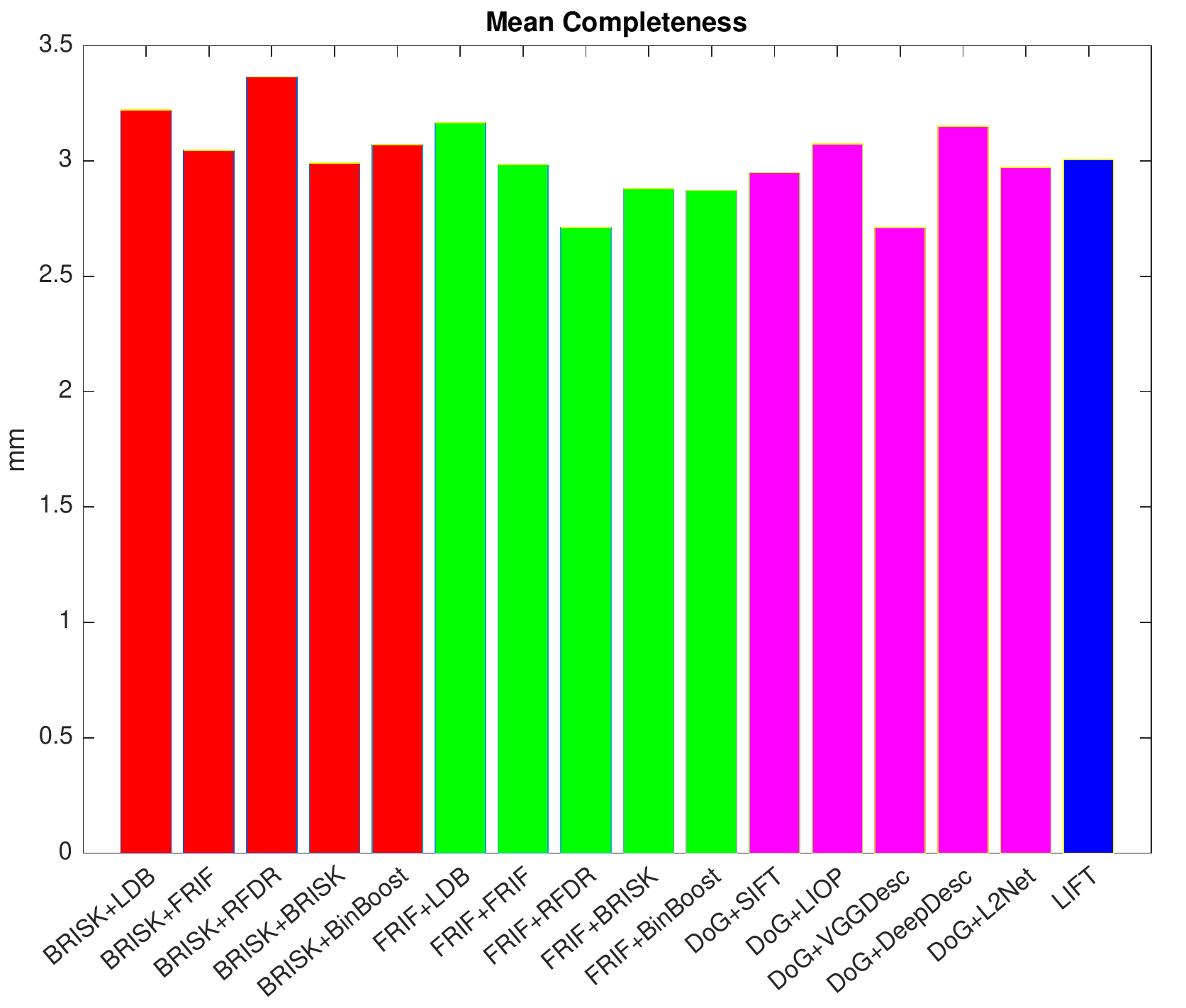}
		\includegraphics[width=0.28\textwidth]{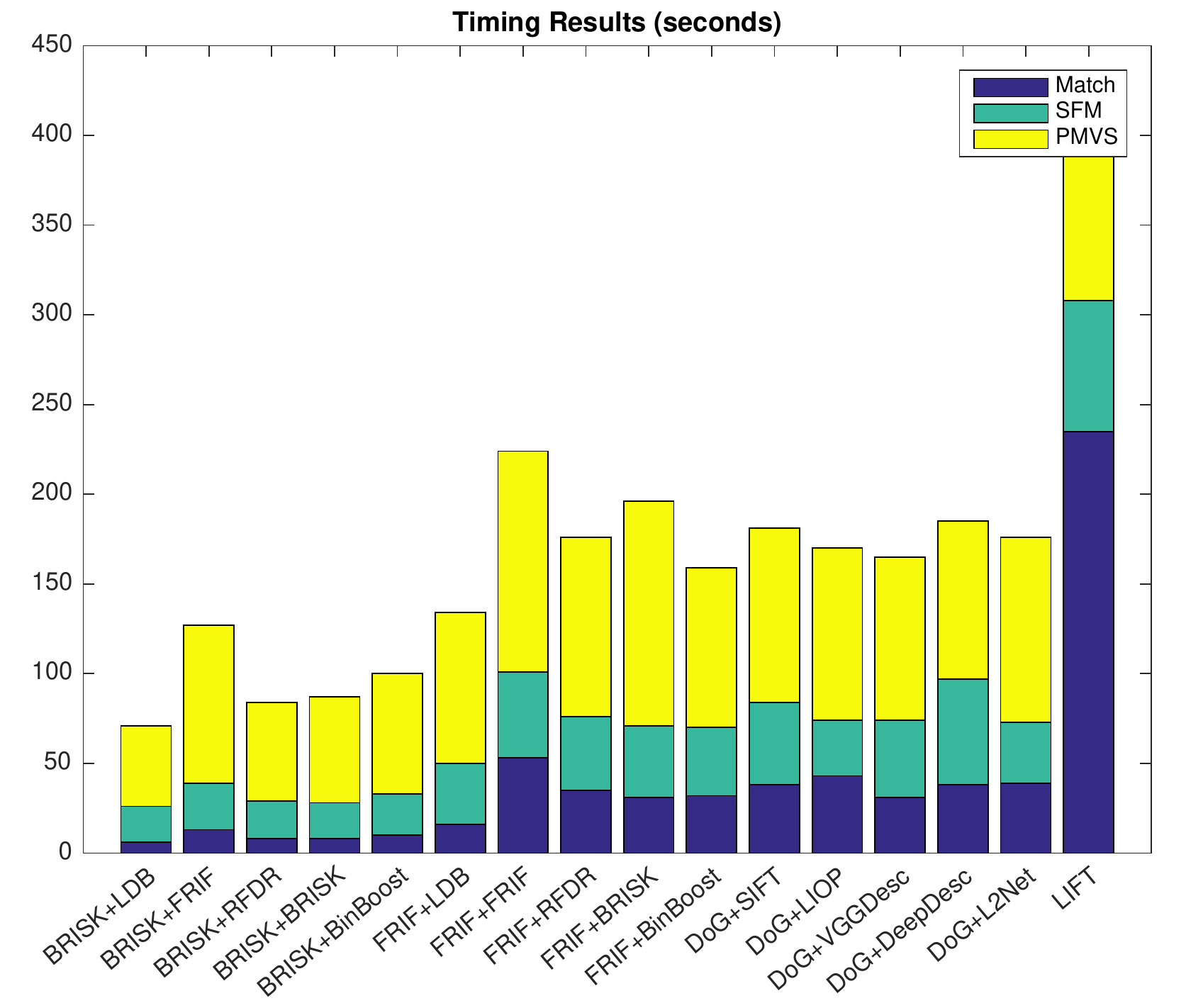}
	\end{minipage}}
		
	\subfloat[]{
	\begin{minipage}[c]{0.12\textwidth}
		\centering
		\includegraphics[width=1.0\textwidth]{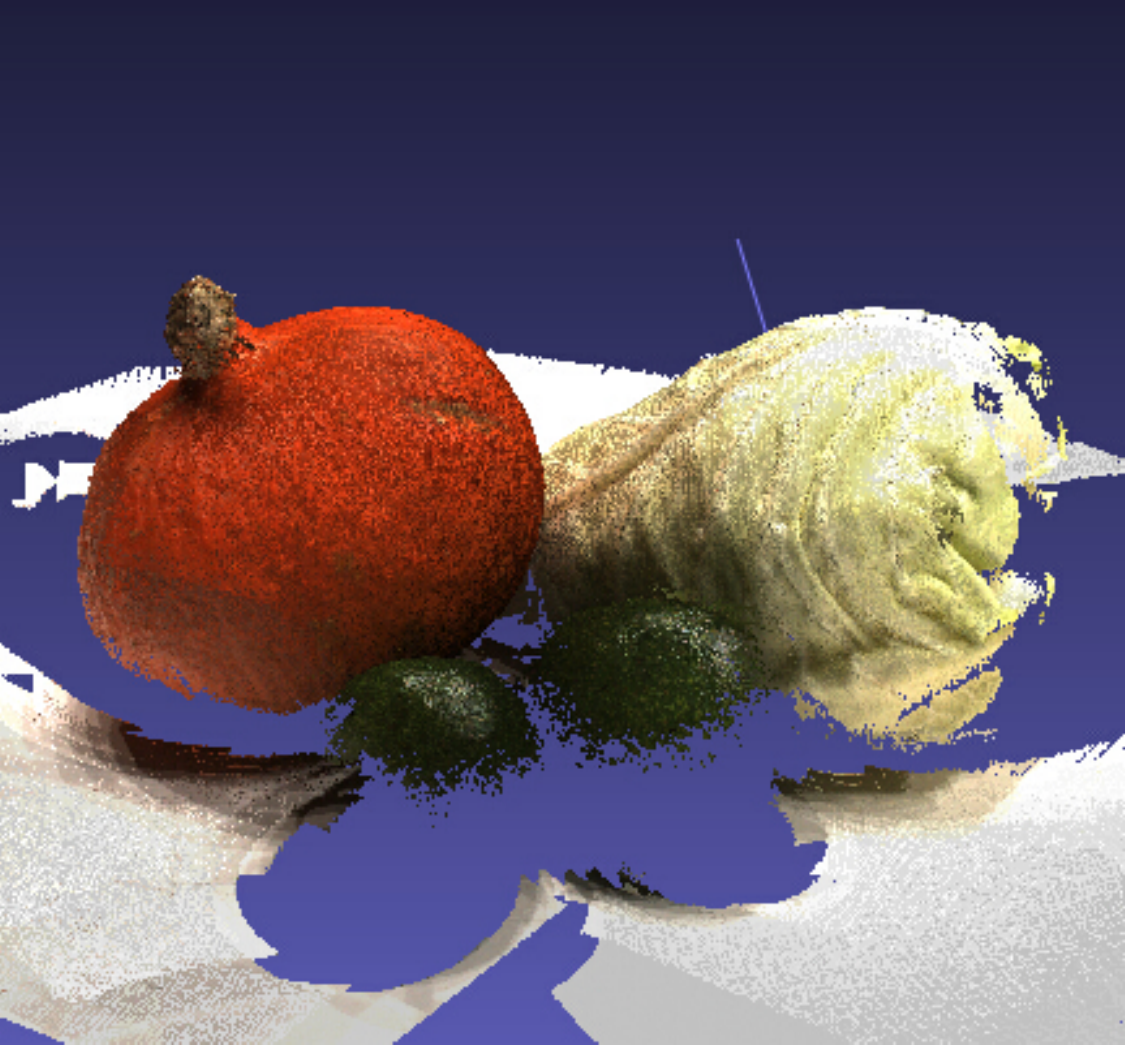}
	\end{minipage}
	\begin{minipage}[c]{0.78\textwidth}
		\centering
		\includegraphics[width=0.28\textwidth]{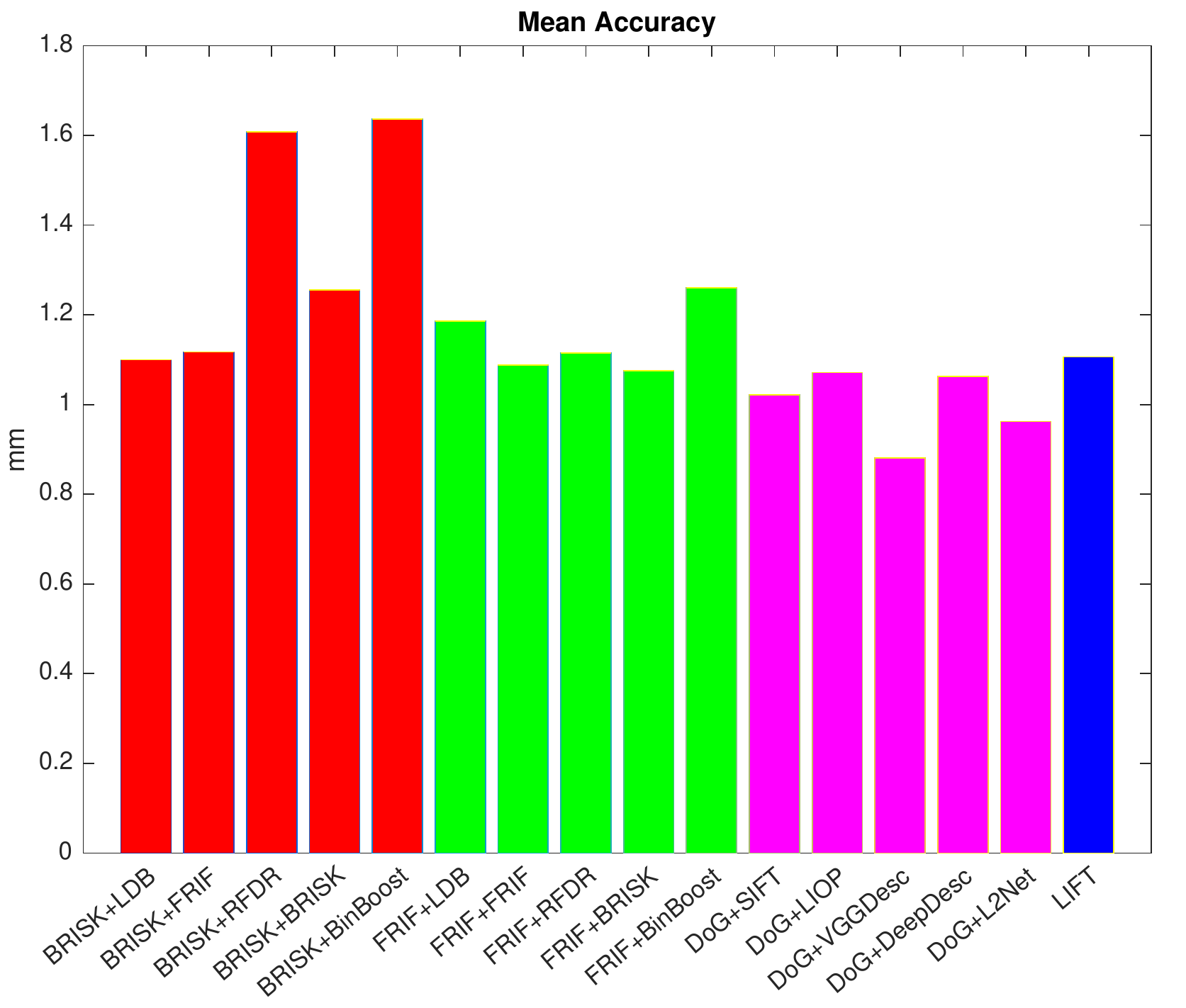}
		\includegraphics[width=0.28\textwidth]{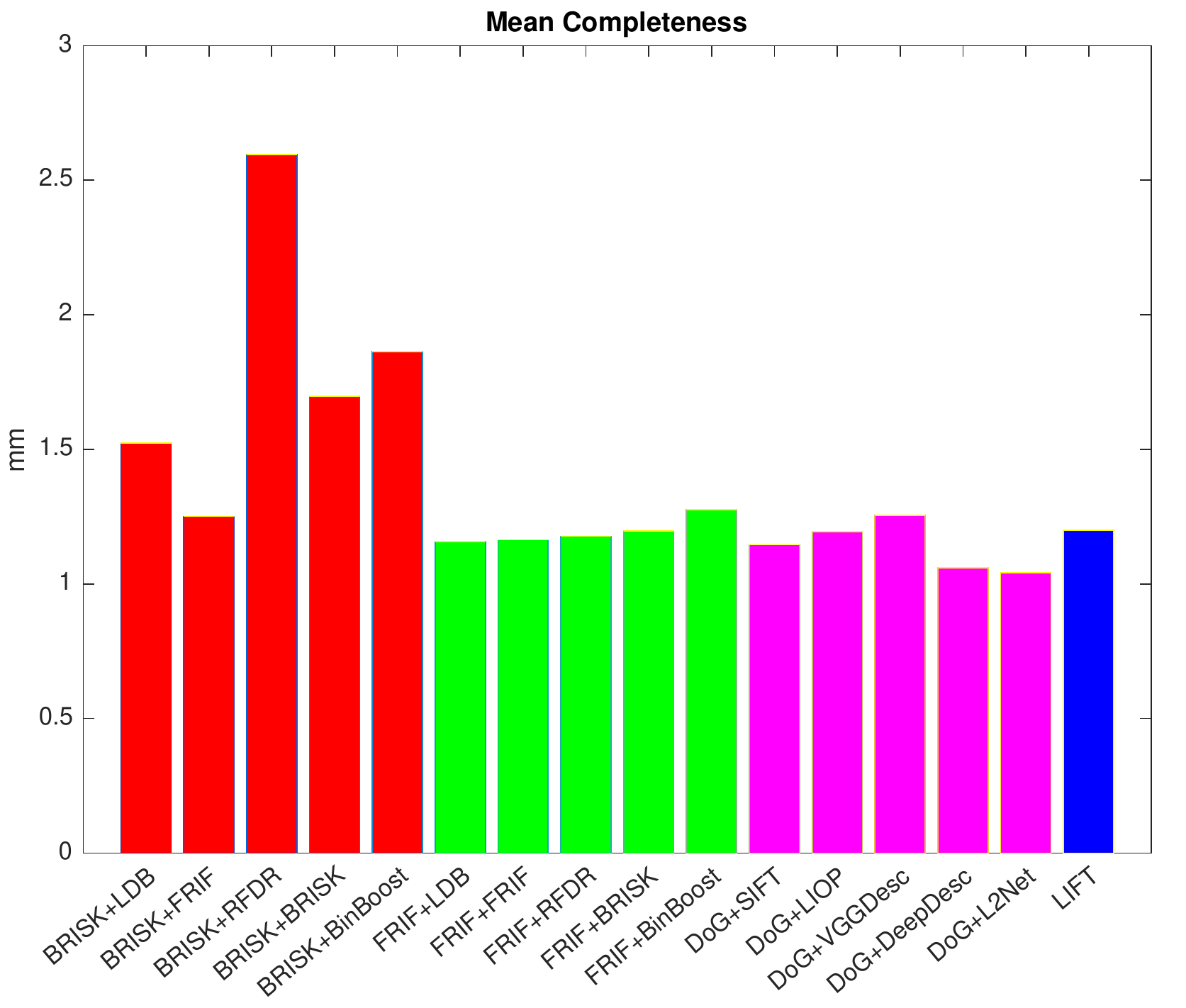}
		\includegraphics[width=0.28\textwidth]{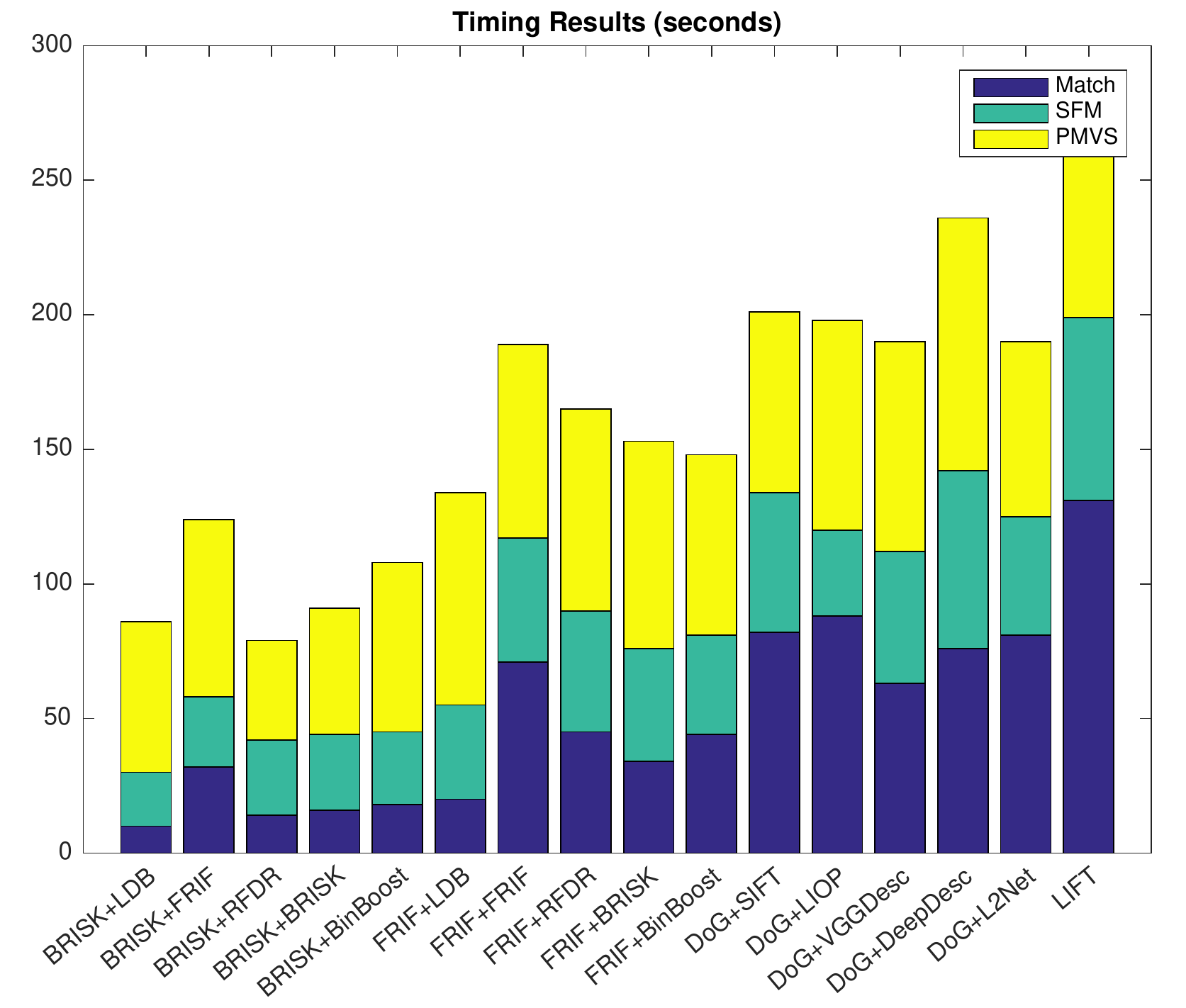}
	\end{minipage}}
   					
	\caption{Performance of scenes that have large accuracy and completeness variances among different evaluated methods. From left to right are: the scene, mean accuracy of different methods, mean completeness of different methods, and the timing results in different stages of different methods. \label{fig:largevar_result}}
\end{figure*}

\textbf{\emph{Scenes with large performance variance}}. In this case, it refers to complex scene types for reconstruction. The results are shown in Fig.~\ref{fig:largevar_result}. In these figures, the 1st column displays the scenes, the 2nd column shows the mean accuracy of different methods, the 3rd column shows the mean completeness of different methods, and the 4th column gives the running times of different methods. From Fig.~\ref{fig:largevar_result}, we have the following observations:

\begin{itemize}
\item Consistent to the observation in easy scenes, using FRIF keypoint is relatively less sensitive to the used descriptors than using the BRISK keypoint. In many scenes, it produces similar results for different binary descriptors when using FRIF keypoint. This property of FRIF is similar to DoG. To further show this point, for each kind of keypoint, we record the number of scenes that have large performance variance for different descriptors. These numbers for BRISK, FRIF and DoG are 7, 3 and 2 respectively.
\item Different from the easy scenes, BRISK with BinBoost does not perform the best for these complex scenes. For these complex scenes, it is hard to say which combination is better because it tends to be scene related. In addition, LIFT does not perform the worst for these complex scenes, but it is the most time consuming. In general, using float features is a better choice than using binary features if one does not consider the running time.
\item For the float type features, the learning based descriptors do not necessarily outperform the handcrafted ones. The baseline SIFT performs rather well for all these complex scenes. Similar results can be observed for the binary features, among which the handcrafted ones are better than many learned ones in most cases.
\item In most cases, the running times of SFM and PMVS for all evaluated methods are similar, the main difference of total running time lies in the matching time. In general, using BRISK keypoint requires less running time than using other keypoints. For either BRISK or FRIF keypoints, using FRIF descriptor requires more matching time than other binary descriptors, thus needs more time to do the reconstruction task. Among all the evaluated methods, using float features is more time consuming since matching binary features is more efficient. Due to the smaller descriptor length, using VGGDesc requires the least running time among all the evaluated float features. L2Net usually requires less time than SIFT and DeepDesc although all of them have the same descriptor length. This implicitly indicates that L2Net could generate better matching results~(i.e., similar number of matches but with higher precision), thus requiring less time for SFM.

\end{itemize}

\begin{figure*}
\centering
    \subfloat[]{
    \begin{minipage}[c]{0.15\textwidth}
        \centering
        \includegraphics[width=1.0\textwidth]{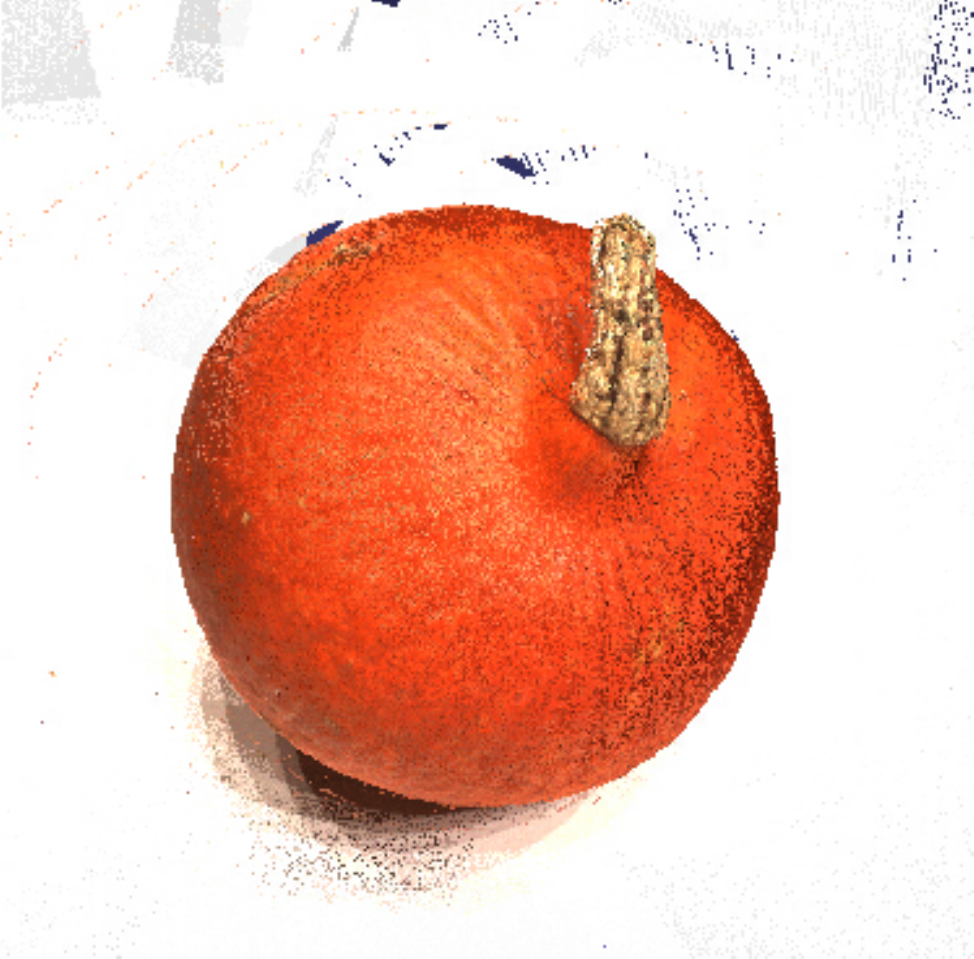}
    \end{minipage}
    \begin{minipage}[c]{0.84\textwidth}
        \centering
        \includegraphics[width=0.32\textwidth]{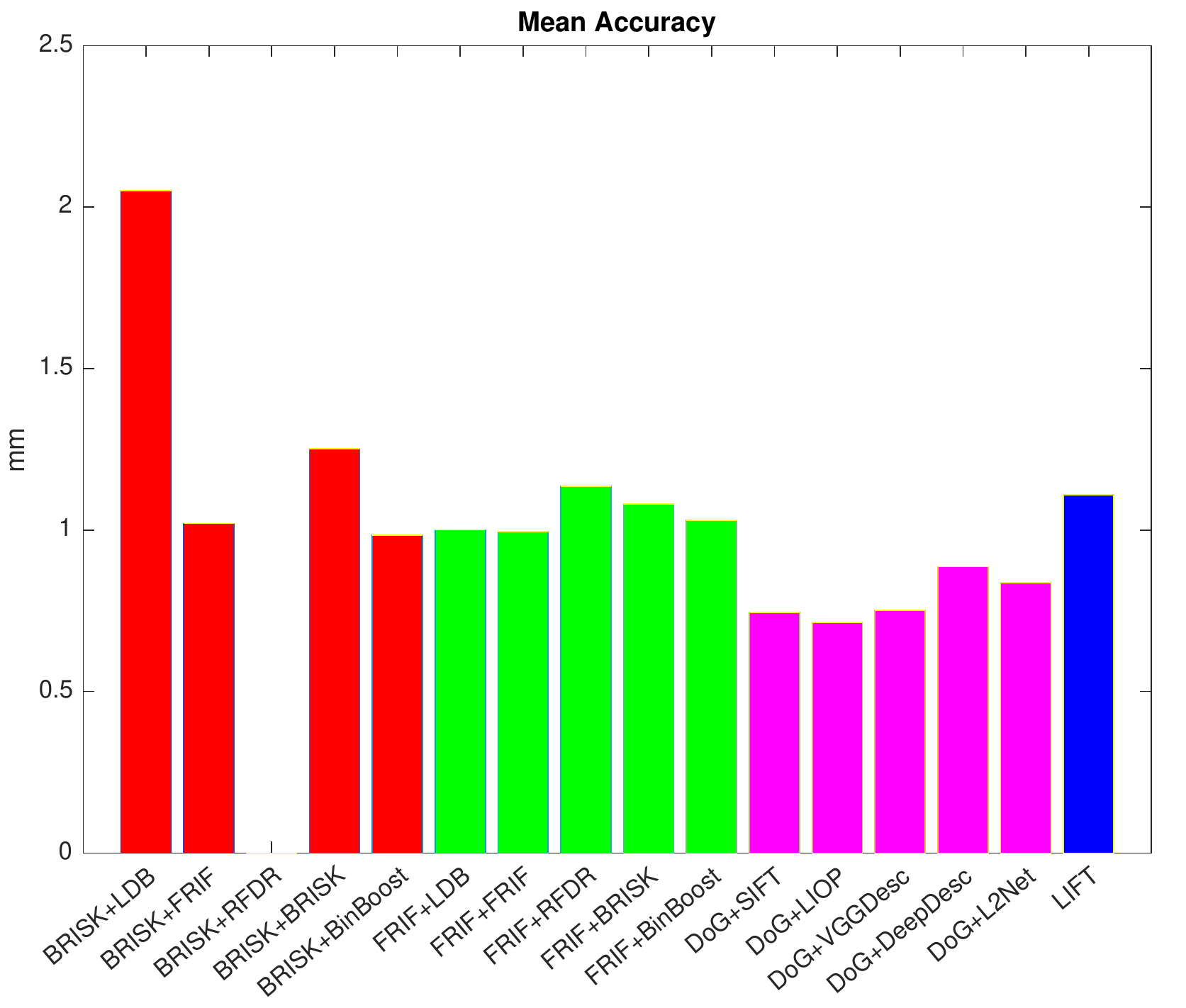}
        \includegraphics[width=0.32\textwidth]{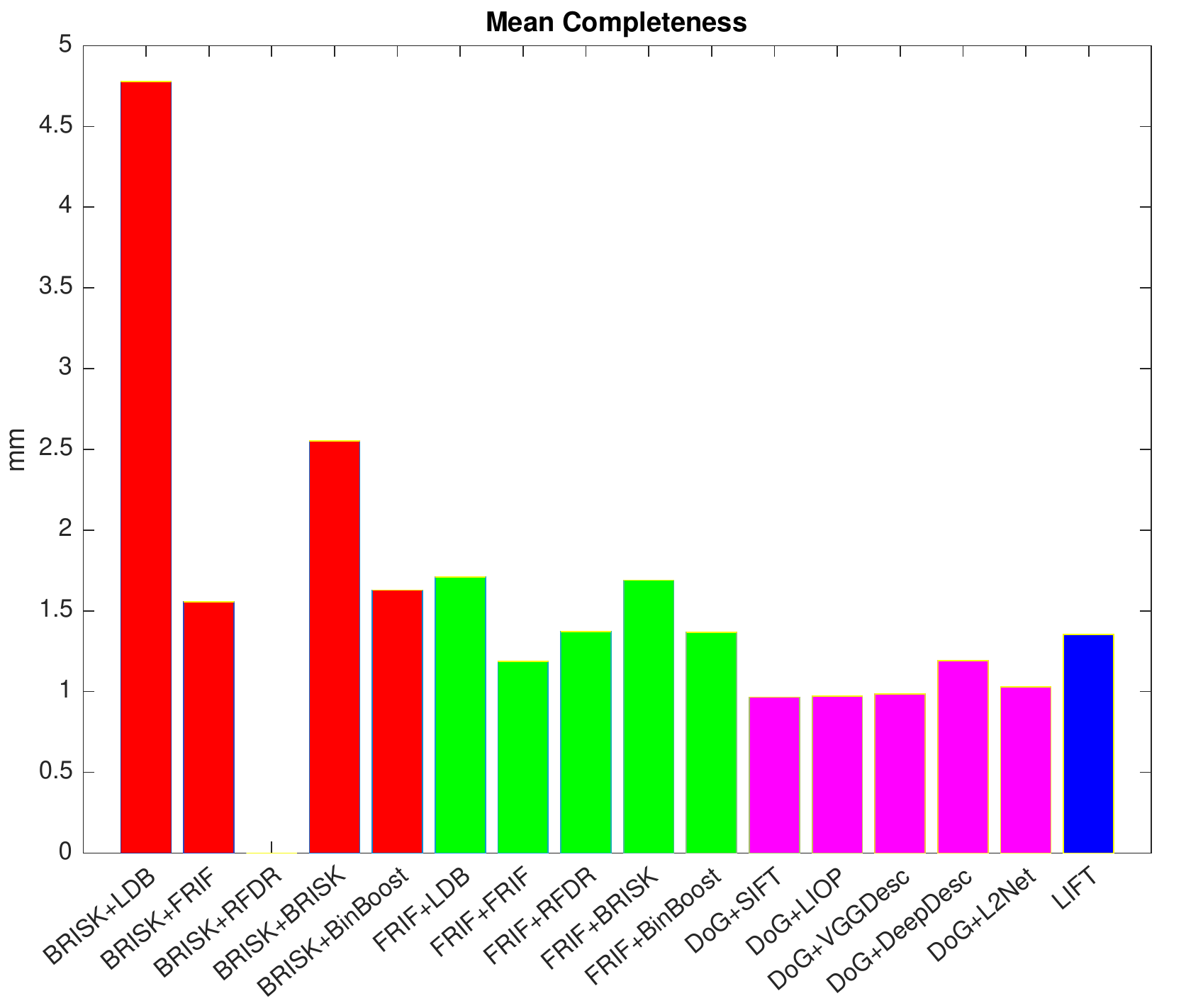}
        \includegraphics[width=0.32\textwidth]{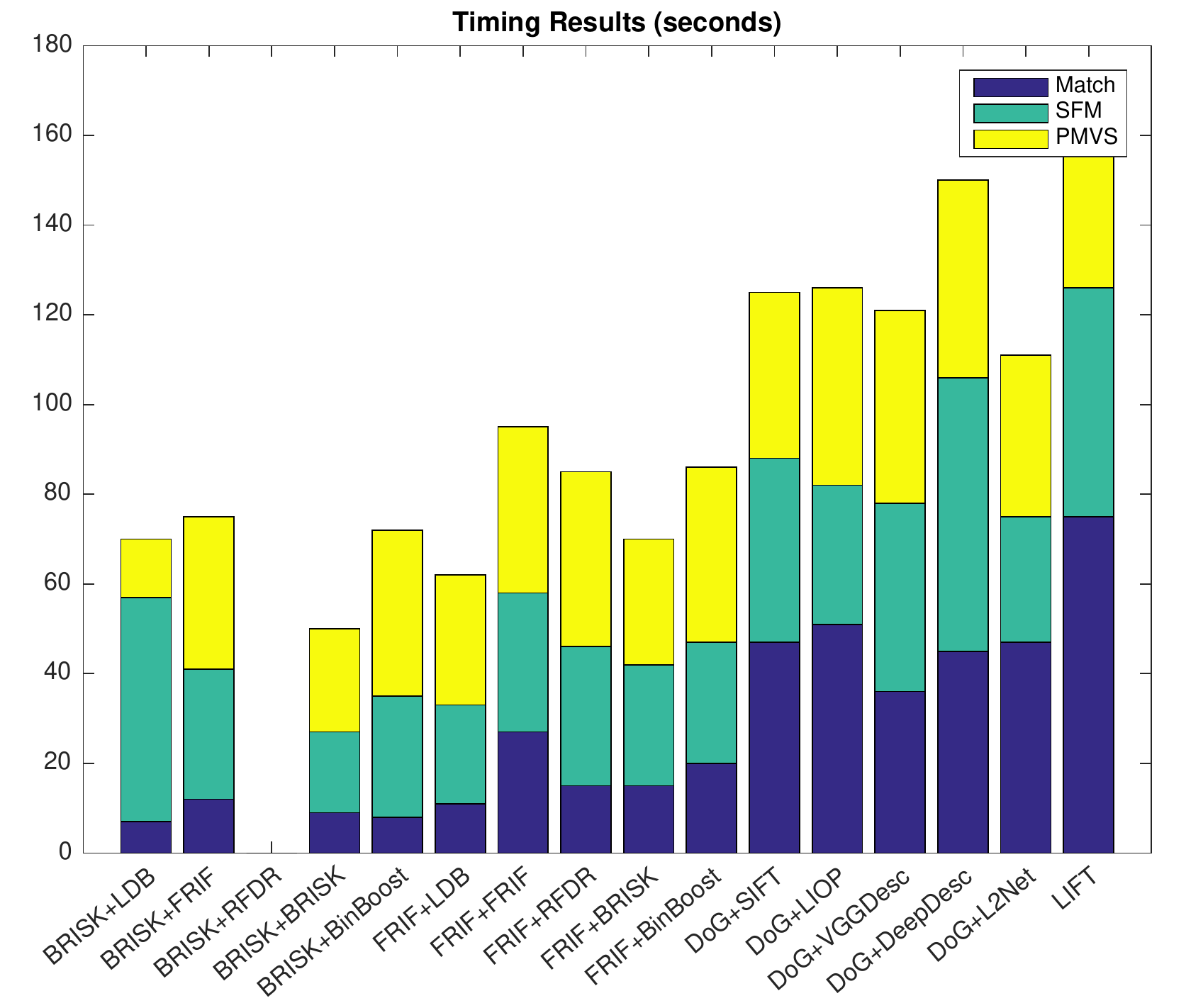}
    \end{minipage}}

    \subfloat[]{
    \begin{minipage}[c]{0.15\textwidth}
        \centering
        \includegraphics[width=1.0\textwidth]{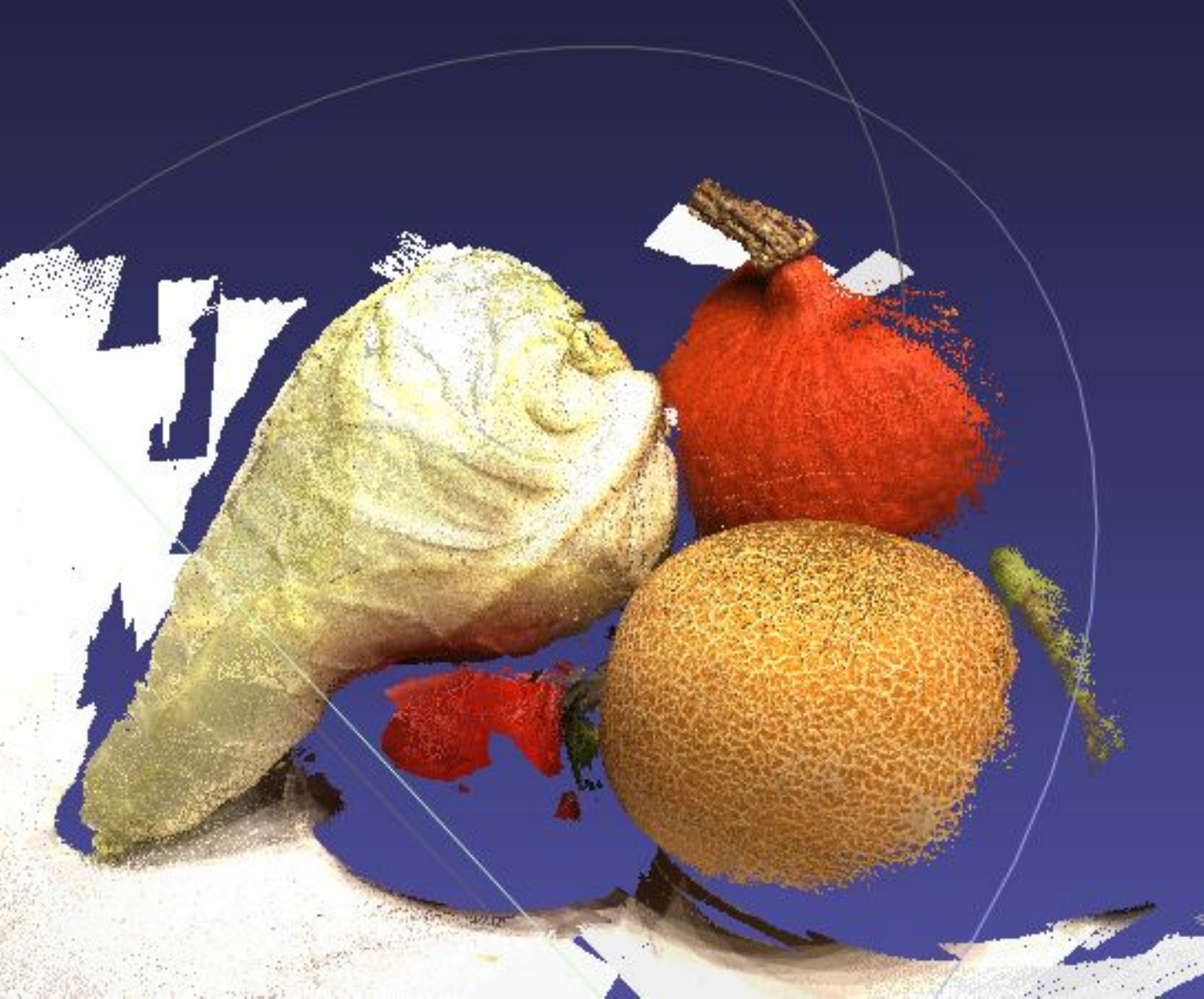}
    \end{minipage}
    \begin{minipage}[c]{0.84\textwidth}
        \centering
        \includegraphics[width=0.32\textwidth]{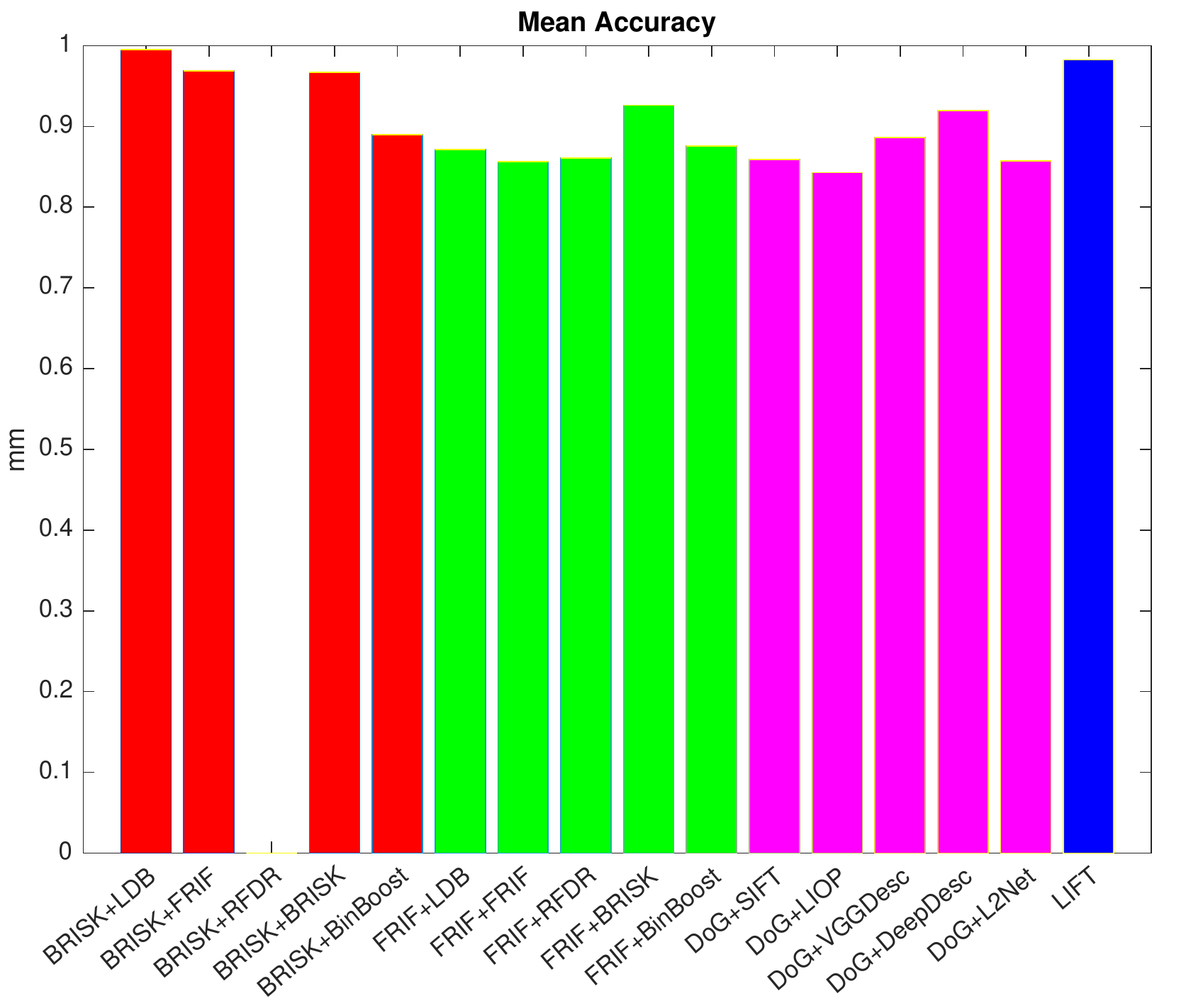}
        \includegraphics[width=0.32\textwidth]{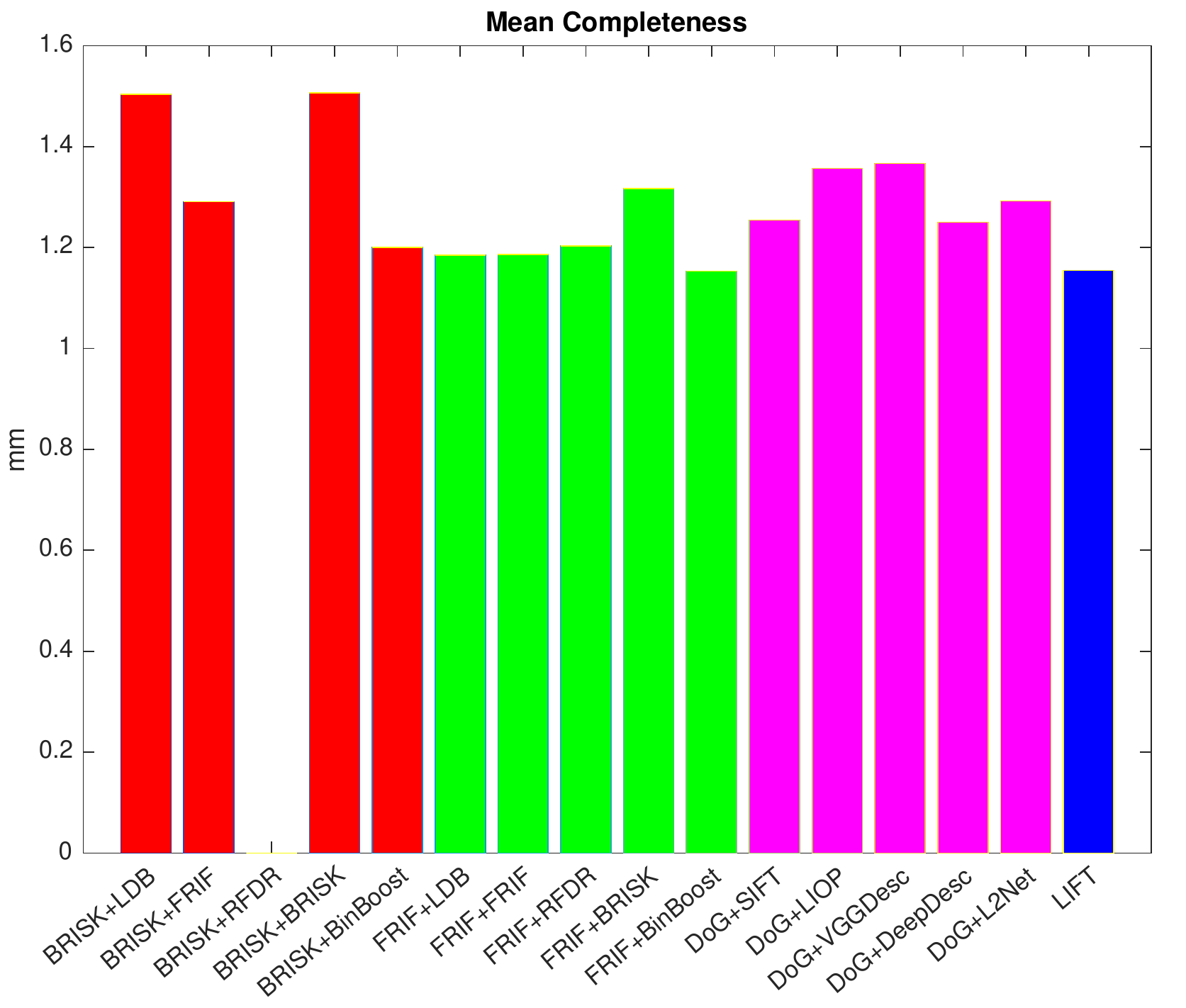}
        \includegraphics[width=0.32\textwidth]{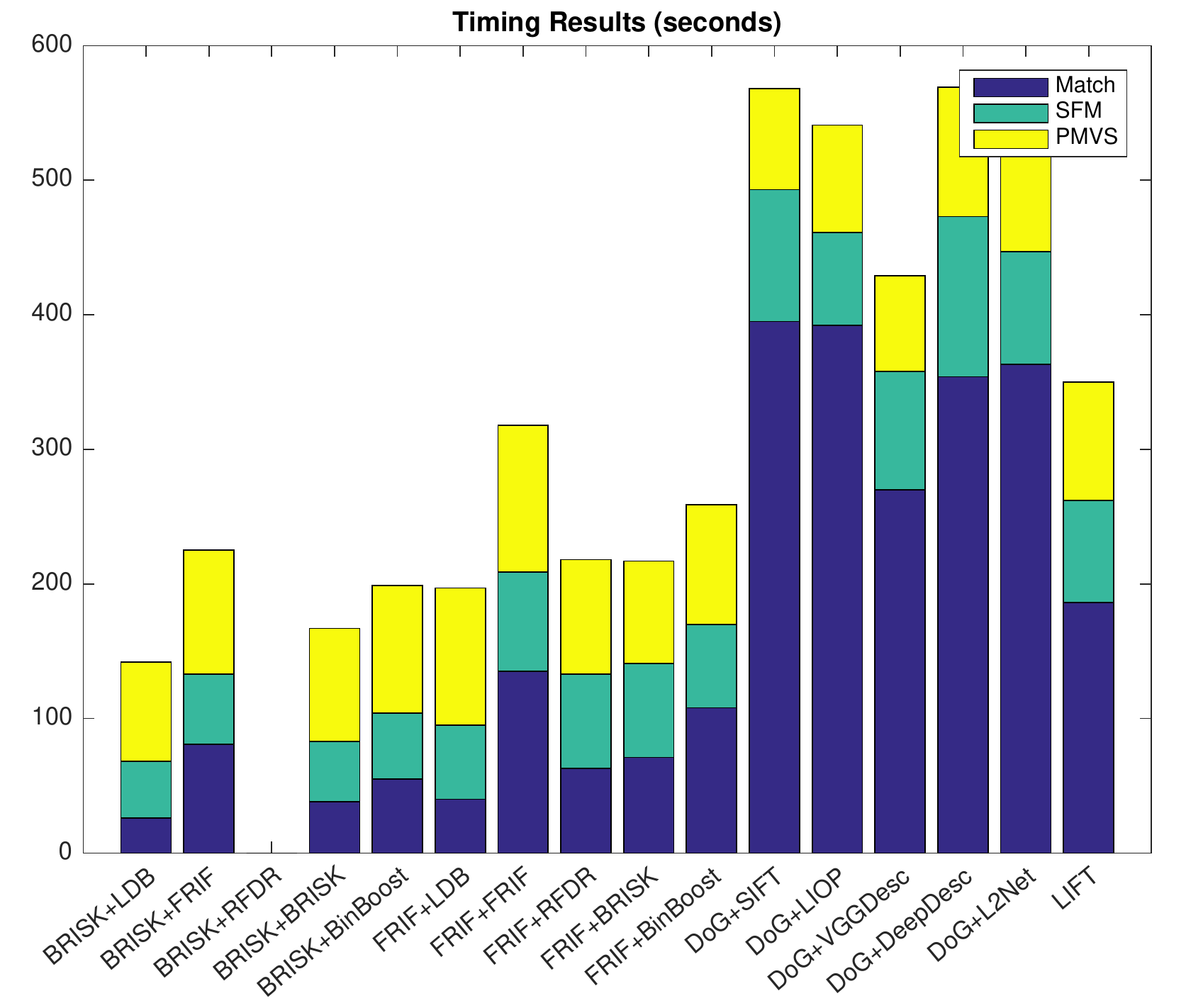}
    \end{minipage}}

    \subfloat[]{
    \begin{minipage}[c]{0.15\textwidth}
        \centering
        \includegraphics[width=1.0\textwidth]{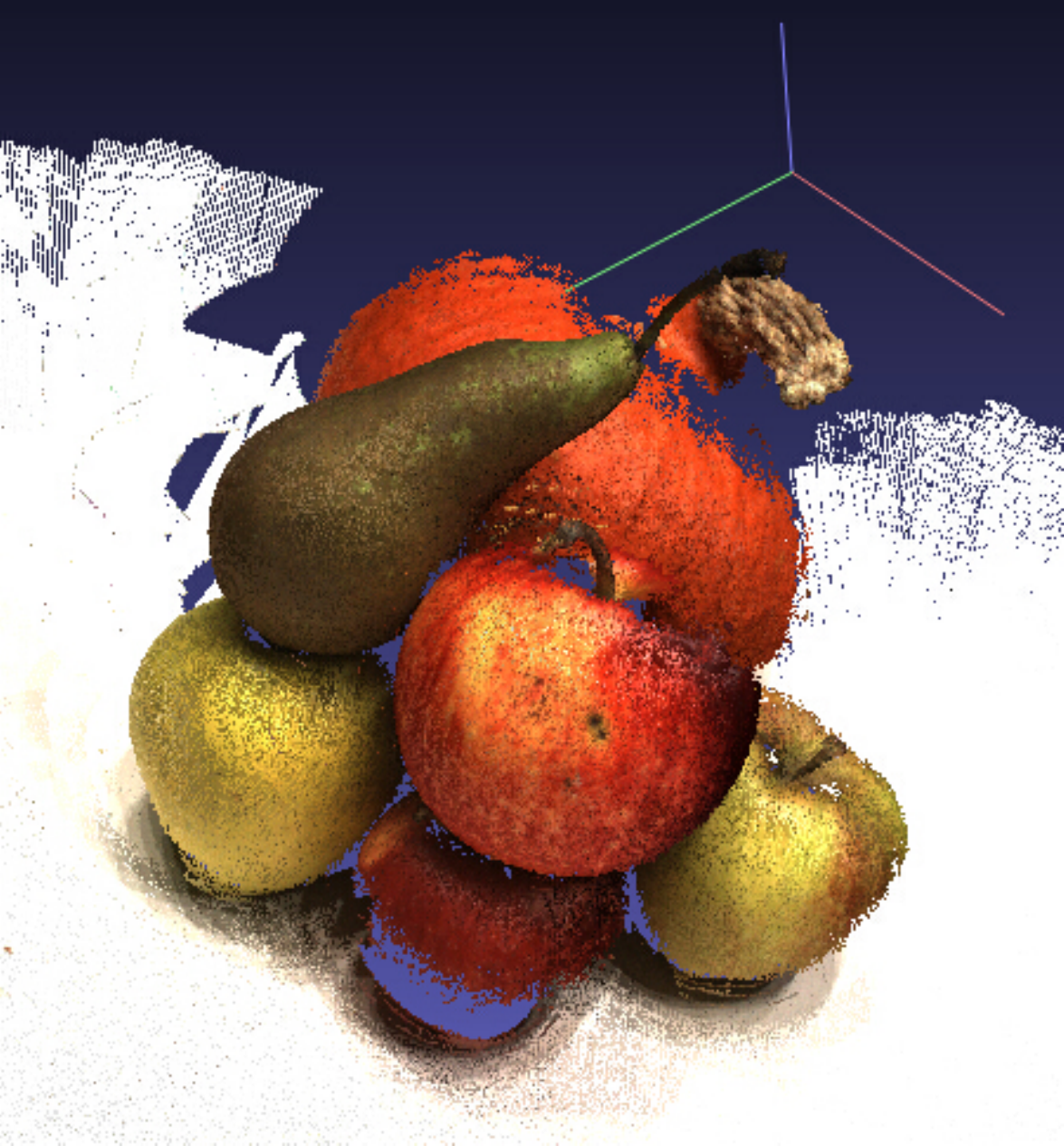}
    \end{minipage}
    \begin{minipage}[c]{0.84\textwidth}
        \centering
        \includegraphics[width=0.32\textwidth]{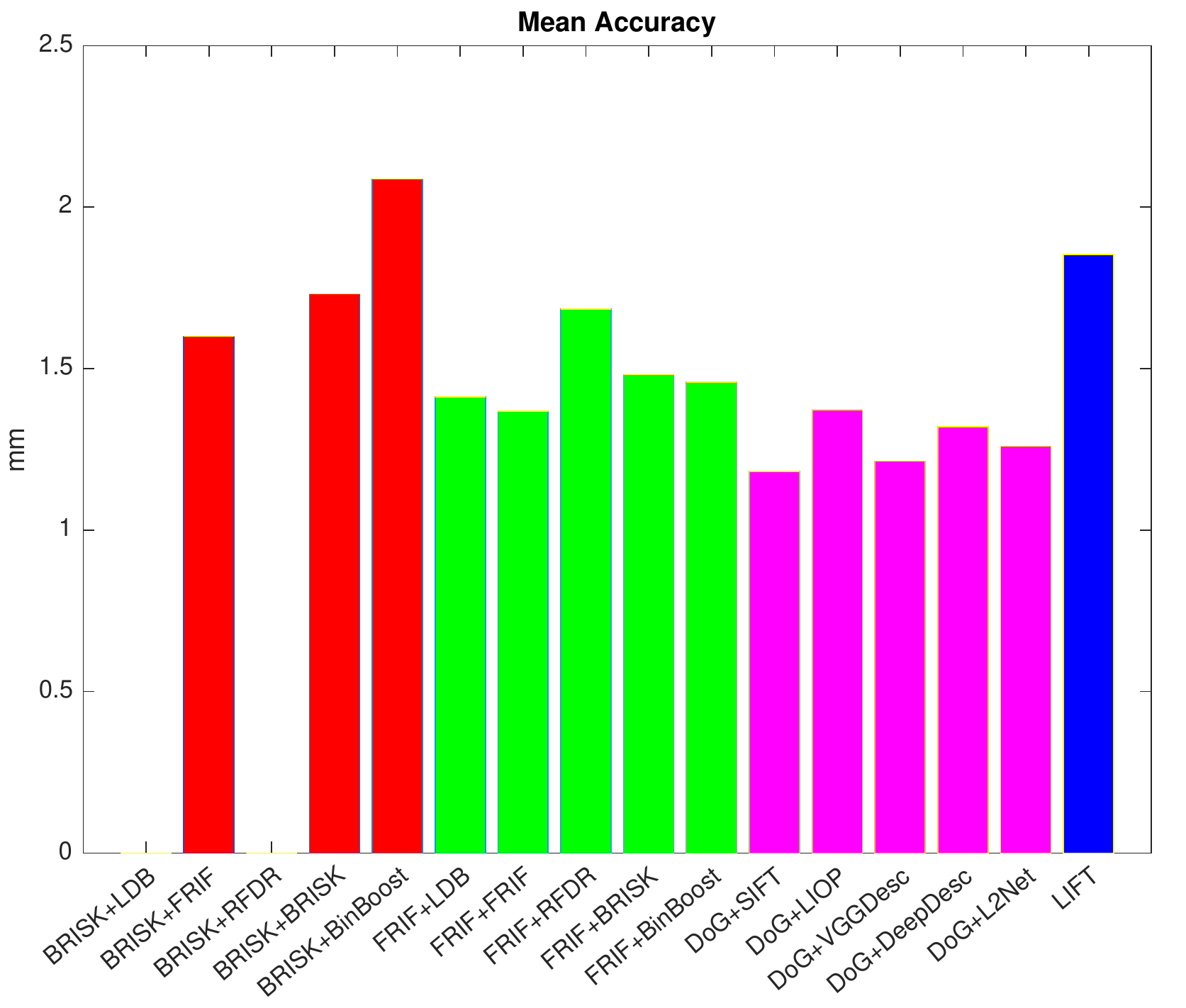}
        \includegraphics[width=0.32\textwidth]{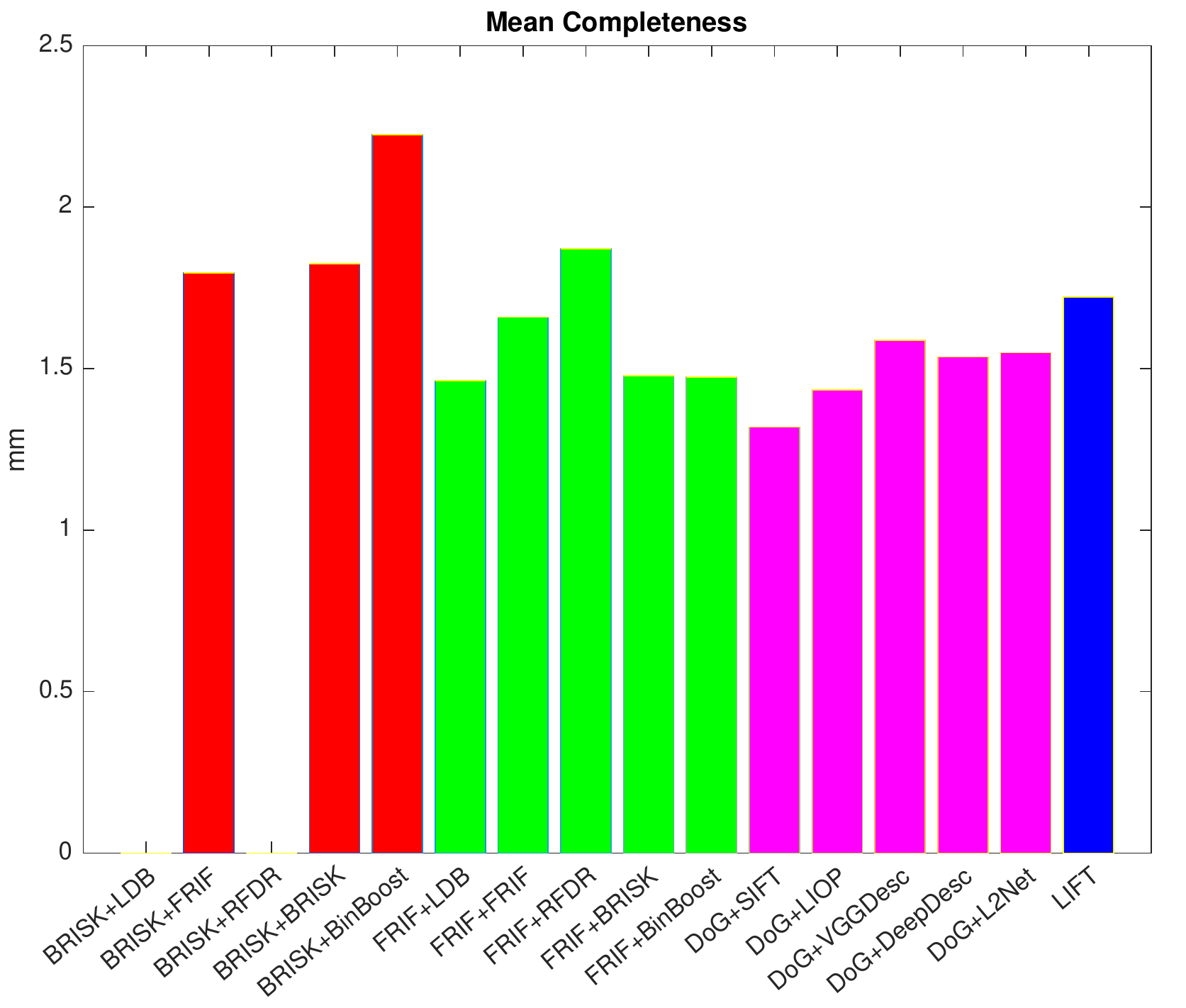}
        \includegraphics[width=0.32\textwidth]{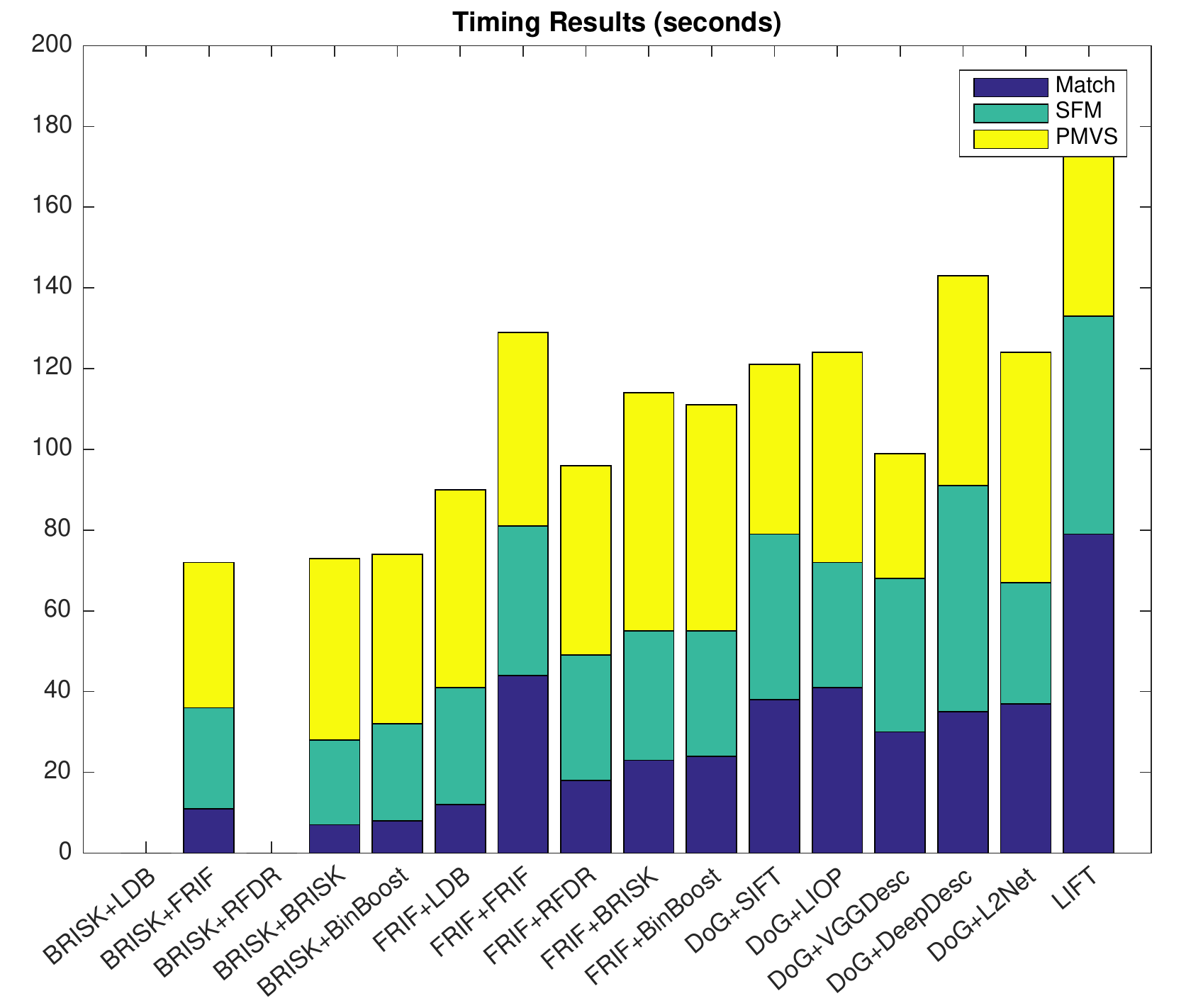}
    \end{minipage}}

	\caption{Performance of scenes that at least one method fails to obtain the reconstruction result. If one method fails, there is no bar shown in the related figures. \label{fig:failscene_result}}
\end{figure*}

\textbf{\emph{Scenes that at least one method fails}}. In this case, it refers to the most challenging scene type for 3D reconstruction since one may fail if the local feature is not chosen appropriately. The results are shown in Fig.~\ref{fig:failscene_result}. We can find that all the failures are from the combinations with BRISK as keypoint detector. More specifically, using LDB descriptor leads to failure for one scene, while using RFD is responsible for 3 failed cases. Even in cases that using BRISK keypoints can be survived to get a reconstruction result, it is usually less accurate and complete than using other keypoints. Considering together with the performance of BRISK keypoint for complex scenes, it is clear that BRISK keypoint is less suitable for reconstructing complex and challenging scene types. However, we have to acknowledge that it is a good choice for easy scene types because it requires less time to obtain better accuracy. While for the other keypoints, DoG is slightly better. Taking Fig.~\ref{fig:smallvar_result} to Fig.~\ref{fig:failscene_result} altogether, it is interestingly to see that when the scene type becomes more and more challenging, using float type features gradually shows its superiority over binary features. Even though, using FRIF keypoint with one binary descriptor is still a good choice for 3D reconstruction with moderate images captured from controlled conditions~(e.g., fixed viewpoints) as it requires less running time than using float type features. While among the float type features, the reconstruction results of LIFT is less accurate due to the larger localization error of LIFT than that of DoG.

\section{Evaluation on Large Scale Structure from Motion Dataset}
\label{sec:SFM_result}
Apart from the controlled case of image capturing, we also evaluate all these local features on 3D reconstruction from a large collection of Internet images, which is the case of most large scale applications of 3D reconstruction, i.e., reconstructing landmarks or cities. For this experiment, we choose the large scale structure from motion dataset~\cite{Wilson_ECCV14}. This dataset contains images of several landmarks across the world. For each landmark, it has several thousands of images obtained from the Internet. Different from the previous tested MVS dataset, each image set of one landmark contains a large portion of unrelated images as distractors. On the contrary, the MVS dataset only contains images of one scene from different viewpoints. Meanwhile, since there is no constraint on these collected images, they inevitably contain many low quality and non-overlapping images. For these reasons, this dataset is more challenging for feature matching, and so for 3D reconstruction.

\begin{figure*}[htbp]
	\centering
    \subfloat[]{
		\includegraphics[width=0.32\textwidth]{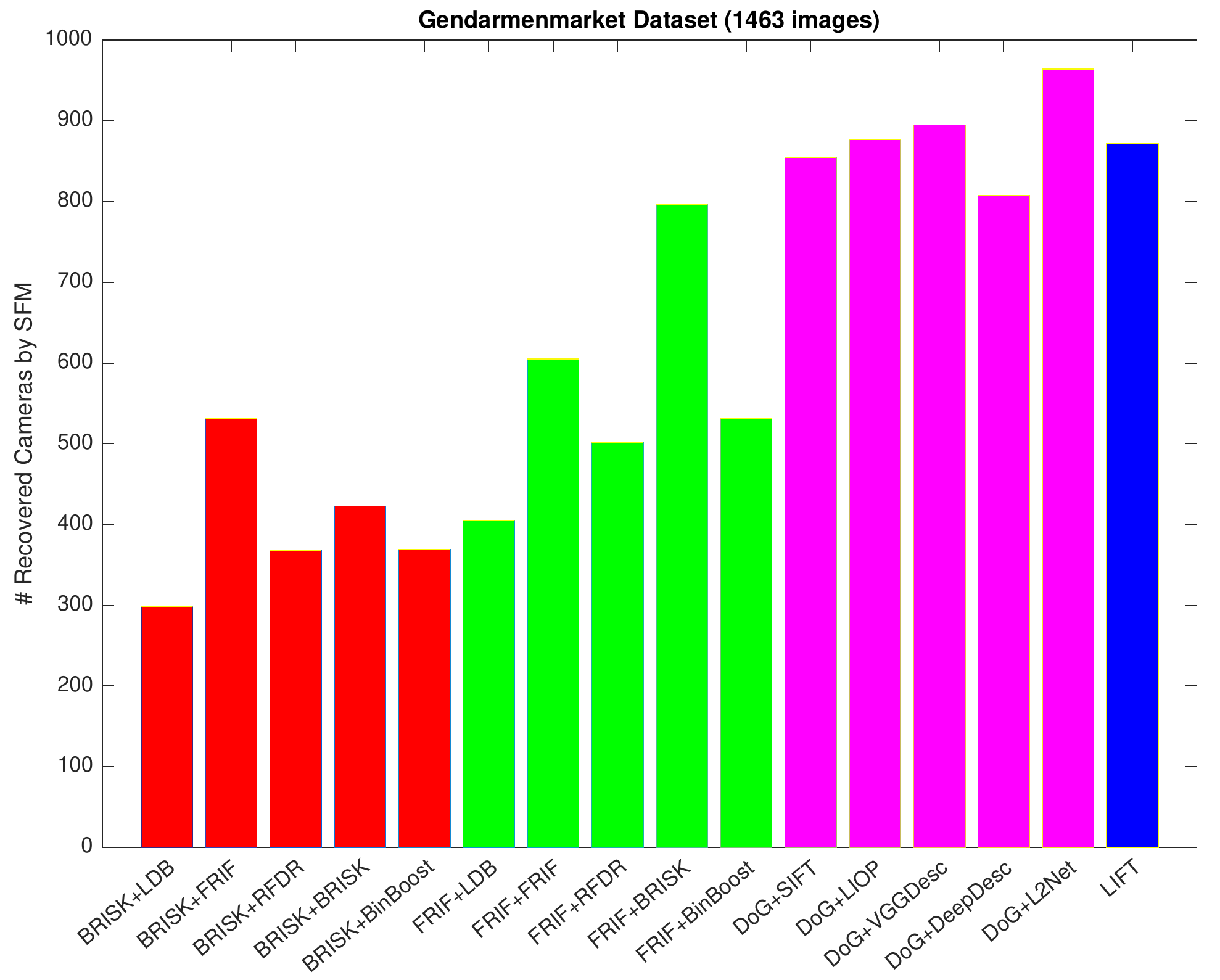}}
	\subfloat[]{
		\includegraphics[width=0.32\textwidth]{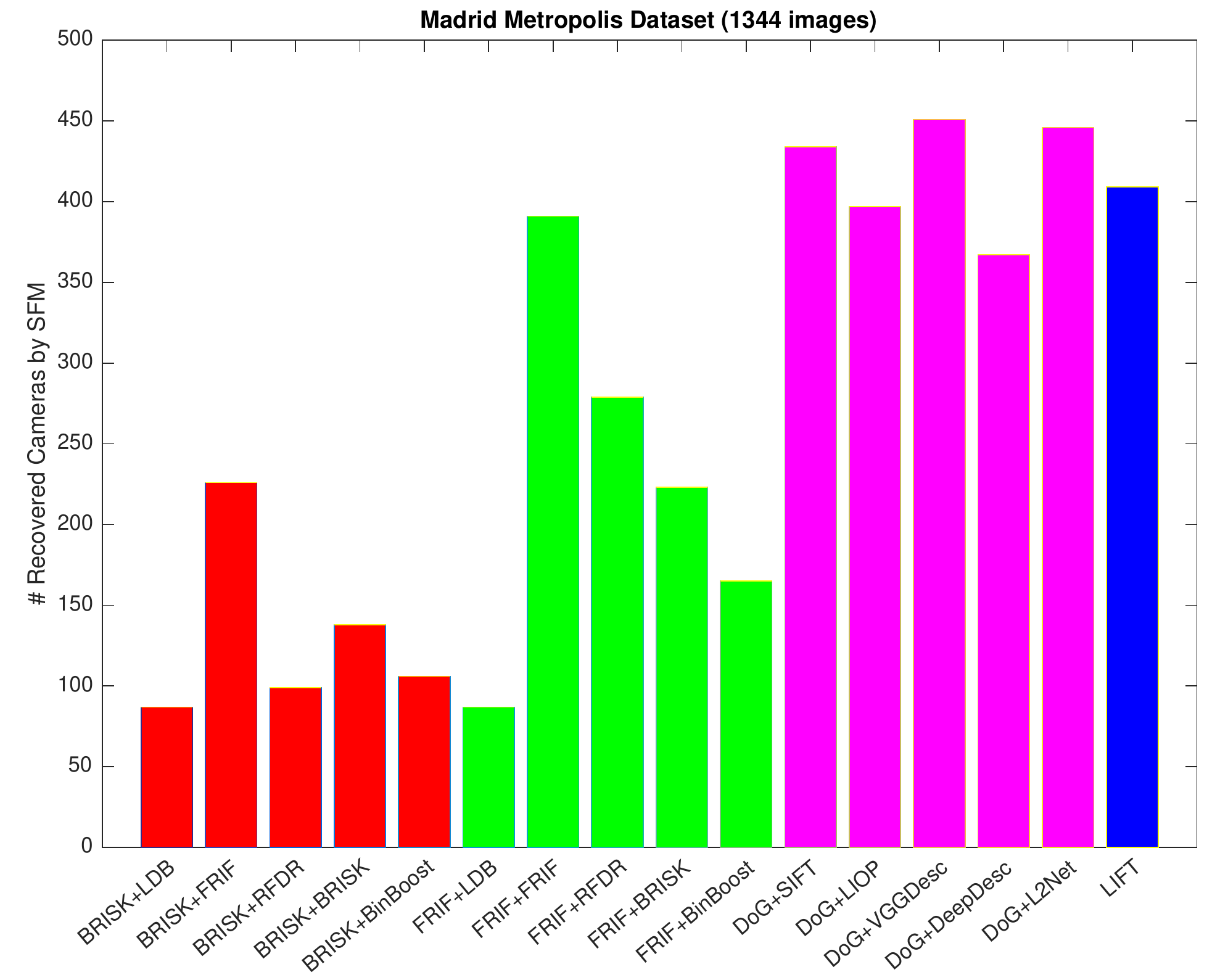}}
    \subfloat[]{
		\includegraphics[width=0.32\textwidth]{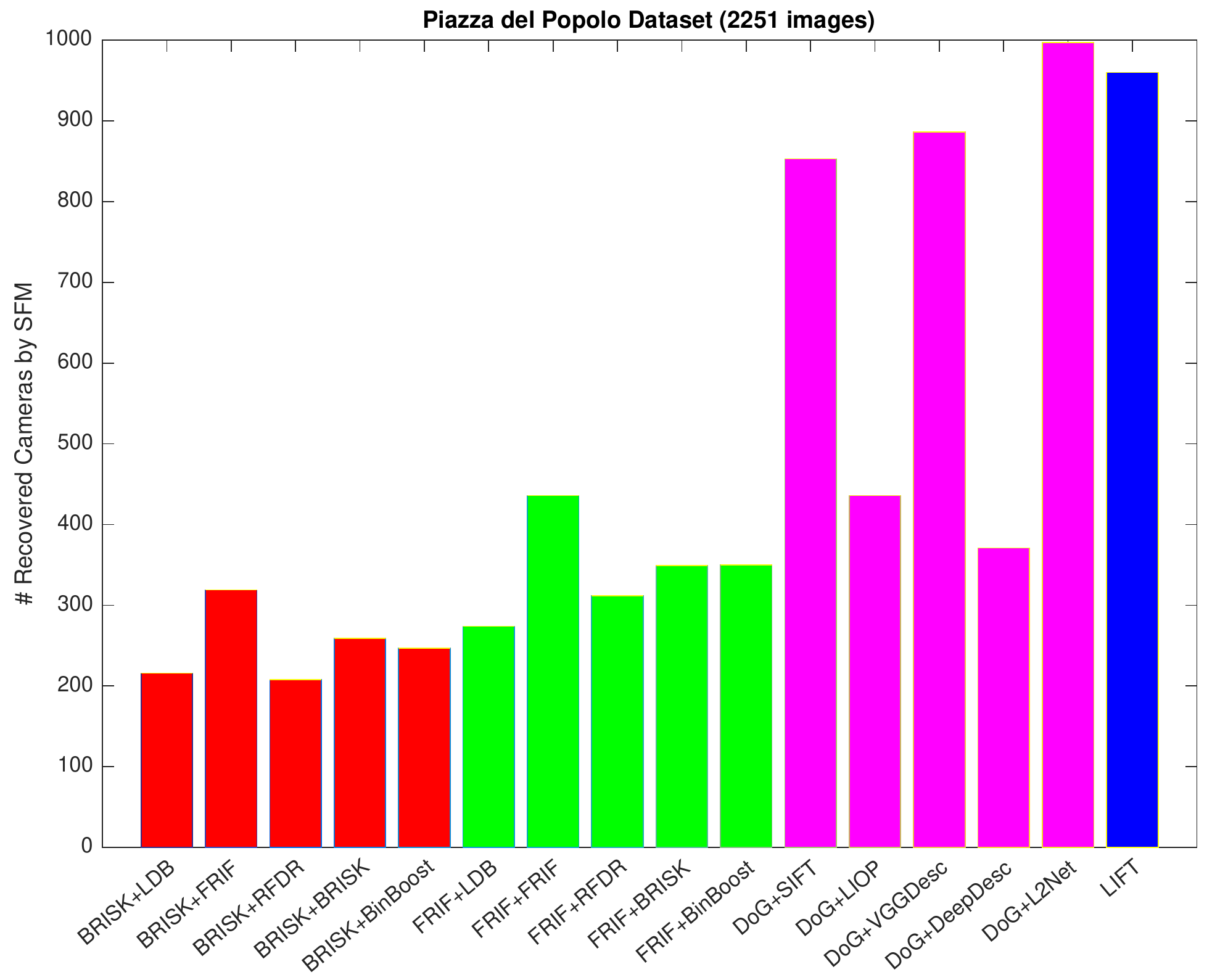}}
	\caption{The number of recovered cameras by SFM based on matching different local features for three different landmarks. \label{fig:SFM_cams}}
\end{figure*}

Since there is no groundtruth 3D model available for this dataset, we use the number of recovered images as the performance indicator for different methods. This is because that the following PMVS procedure is highly related to the number of recovered cameras. In general, if we could recover more number of cameras, the reconstruction could cover more parts of the scene, so the more number of 3D points could be obtained by PMVS and a better accuracy and completeness are expected for the reconstructed scene. The results are shown in Fig.~\ref{fig:SFM_cams}. In this dataset, the float type features generally perform better than the binary ones with a significantly large margin. This observation is different from the one observed in the previous MVS dataset, where using binary features could achieve comparable results to those of using float type features. Such a superior performance of the float type features demonstrates their good generalization ability. Considering the fact that there are many unrelated images exist in this dataset, binary features may be sensitive to the distractors, i.e., the local features extracted from unrelated images. For the binary features, using FRIF keypoint recovers more number of cameras than using BRISK keypoint. In some case, when combined with an appropriate descriptor, using FRIF keypoint can even produce comparable performance to that of using float type features. The better result of using FRIF keypoint than using BRISK keypoint is also consistent to the observations found in MVS dataset. For the handcrafted float type descriptors, the performance of the most traditional SIFT is very stable across different landmarks while LIOP fails to reconstruct a large part of the scene for the third landmark~(Fig.~\ref{fig:SFM_cams}(c)). This is similar for the DeepDesc, showing an inferior performance to other learning based methods. Especially, the advanced CNN based learning method, L2Net, performs the best, which is followed by the traditional learning method, i.e., convex optimization. Both of them outperform the SIFT baseline. It is worth to note that LIFT recovers many cameras for this dataset, implying a potential good performance. This is not contradictory to its inferior performance in reconstruction accuracy shown in the MVS dataset. The reason is that the localization precision of LIFT keypoints is not as accurate as other handcrafted keypoints, but the LIFT descriptor does have a very good matching ability. Therefore, it could recover many cameras but with a relative large error on the recovered camera poses, which would further reduce the reconstruction accuracy as shown in the previous experiments.

\section{Conclusion}
\label{sec:conclusion}
In this paper, we provide an extensively comparative study of popular local features for the task of 3D reconstruction. We focus on how the matching quality of different local features affects the final reconstruction performance, either in terms of accuracy and completeness or indicated by the number of recovered cameras. Our evaluation covers a wide range of the state of the art local features, ranging from the traditional handcrafted ones to the recently popular learning based ones. Meanwhile, we also include both float type feature descriptors and binary ones to have a thorough and comprehensive evaluation. Not only the studied local features have a large diversity, the evaluated datasets also cover the two main application situations of image based 3D reconstruction. One is a controlled case where all images are taken from different viewpoints of the reconstructed scene so that all images have a considerable range of overlap. The other is a general case where many unrelated images exist in the image set of the reconstructed scene. For the first case, we choose to use the recently proposed DTU MVS datasets, which contain various scene types with specifically designed image capturing positions and supply the groundtruth 3D points that facilitate an objective and quantitative evaluation of the reconstruction results. While for the latter case, we choose to use the Internet scale image sets of landmarks, each of which contains a large number of related images and distractors.

Such a dedicated consideration on the evaluated methods and datasets makes our work potentially be a guidance for practical engineers on 3D reconstruction applications. Our experimental results reveal that for the controlled case where no distracting images exist, using binary features is good enough to produce the state of the art 3D reconstruction results with only a fraction of time of using float type features. However, for the large scale free image set with many distractors, using binary features can not guarantee the good performance. The float type descriptors are the most competitive ones in this case even though they need more time to establish point correspondences. Among the evaluated float type descriptors, using recently learned descriptors, such as VGGDesc and L2Net, can lead to better results than using handcrafted ones~(SIFT, LIOP). However, DeepDesc is not as competitive as these two learned descriptors. Meanwhile, the most traditional SIFT also produces very good results among all the evaluated features, implying that it still requires a lot of efforts to improve the general matching performance of local features. The good results of the learned descriptors further encourage the potential of learning descriptors. However, how to learn the whole stuffs of feature detection and description together still requires lots of works to do, as shown by the results of LIFT which are even inferior to the baseline in terms of reconstruction accuracy and completeness, indicating a less accurate localization of the learned keypoints.

%\section*{Acknowledgment}
%This work was supported in part by the National Science Foundation of China~.

%\bibliographystyle{IEEEtran}
%\bibliography{journal,conference}

\end{document}